\documentclass[letterpaper, 10 pt, conference]{ieeeconf}  

\IEEEoverridecommandlockouts                              

\usepackage{amsmath}
\usepackage{bm}
\usepackage{amsfonts}
\usepackage{multirow}
\usepackage{graphicx}
\usepackage{subfig}
\usepackage{tikz}
\usepackage{cite}
\usepackage{subfig}

\usepackage{graphicx}
\usepackage{mwe}
\let\labelindent\relax
\usepackage{enumitem}
\usepackage{adjustbox}
\usepackage[most]{tcolorbox}

\usepackage{algorithm2e}
\usepackage[font=footnotesize]{caption}
\usepackage[noend]{algpseudocode}
\usepackage{hyperref,cleveref}
\usepackage[colorinlistoftodos,prependcaption,textsize=tiny]{todonotes}
\usepackage{acronym}
\usepackage{pgfplots}

\newlength\Fcolumnseprule
\usepackage[colorinlistoftodos]{todonotes}

\acrodef{CFAR}		[CFAR]			{Constant False-Alarm Rate}
\acrodef{FMCW}		[FMCW]			{Frequency-Modulated Continuous Wave}
\acrodef{TITLE}		[CFEAR Radarodometry]			{Conservative Filtering for Efficient and Accurate Radar Odometry}

\newcommand{\zmin}{z_\mathrm{min}}

\title{\LARGE \bf
CFEAR Radarodometry - Conservative Filtering for Efficient and Accurate Radar Odometry 
}

\author{Daniel Adolfsson, Martin Magnusson, Anas Alhashimi, Achim J. Lilienthal, Henrik Andreasson 
  \thanks{The authors are with the MRO lab of the AASS research centre
    at \"Orebro University, Sweden.
    E-mail: \texttt{Daniel.Adolfsson@oru.se}}%
  \thanks{This work has received funding from the Swedish Knowledge Foundation (KKS) project ``Semantic Robots'' and European Union's
    Horizon~2020 research and innovation programme under grant
    agreement No 732737 (ILIAD).}%
}

\begin{document}

\maketitle
\thispagestyle{empty}
\pagestyle{empty}

\begin{abstract}

This paper presents the accurate, highly efficient, and learning-free method CFEAR Radarodometry for large-scale radar odometry estimation. By using a filtering technique that keeps the \textit{k strongest} returns per azimuth and by additionally filtering the radar data in Cartesian space, we are able to compute a sparse set of oriented surface points for efficient and accurate scan matching. Registration is carried out by minimizing a point-to-line metric and robustness to outliers is achieved using a Huber loss. We were able to additionally reduce drift by jointly registering the latest scan to a history of keyframes and found that our odometry method generalizes to different sensor models and datasets without changing a single parameter. We evaluate our method in three widely different environments and demonstrate an improvement over spatially cross-validated state-of-the-art with an overall translation error of 1.76\% in a public urban radar odometry benchmark, running  at 55~Hz merely on a single laptop CPU thread.

\end{abstract}

\section{Introduction}
Estimating odometry in large-scale GNSS-denied environments is an essential part in any robust autonomous driving or robotics system. Popular systems are based on lidar~\cite{todor_ndt,gicp,loam,behley2018rss}, vision~\cite{zhu2017image,Buczko_vis_odom,6906584}, introceptive sensors such as IMU or wheel odometry, or a combination of various sensors~\cite{7139486,8202318,narula_all-weather_2020}.

Lidar is an accurate and fast long-range sensor but struggles to perceive the environment in presence of dust or in harsh weather~\cite{acarballo2020libre}. Vision is moderately accurate and inexpensive but operates at shorter range and is sensitive to illumination changes or weather conditions. In contrast, radar is a long-range sensing modality that operates well in all weather conditions and is hardly affected by smoke and dust. As radars have recently become compact enough for autonomous systems, it is desired to utilize their excellent properties for robust localization.
\begin{figure}
      {\includegraphics[trim={0cm 0cm 0cm 0cm},clip,width=0.99\hsize,angle=0]{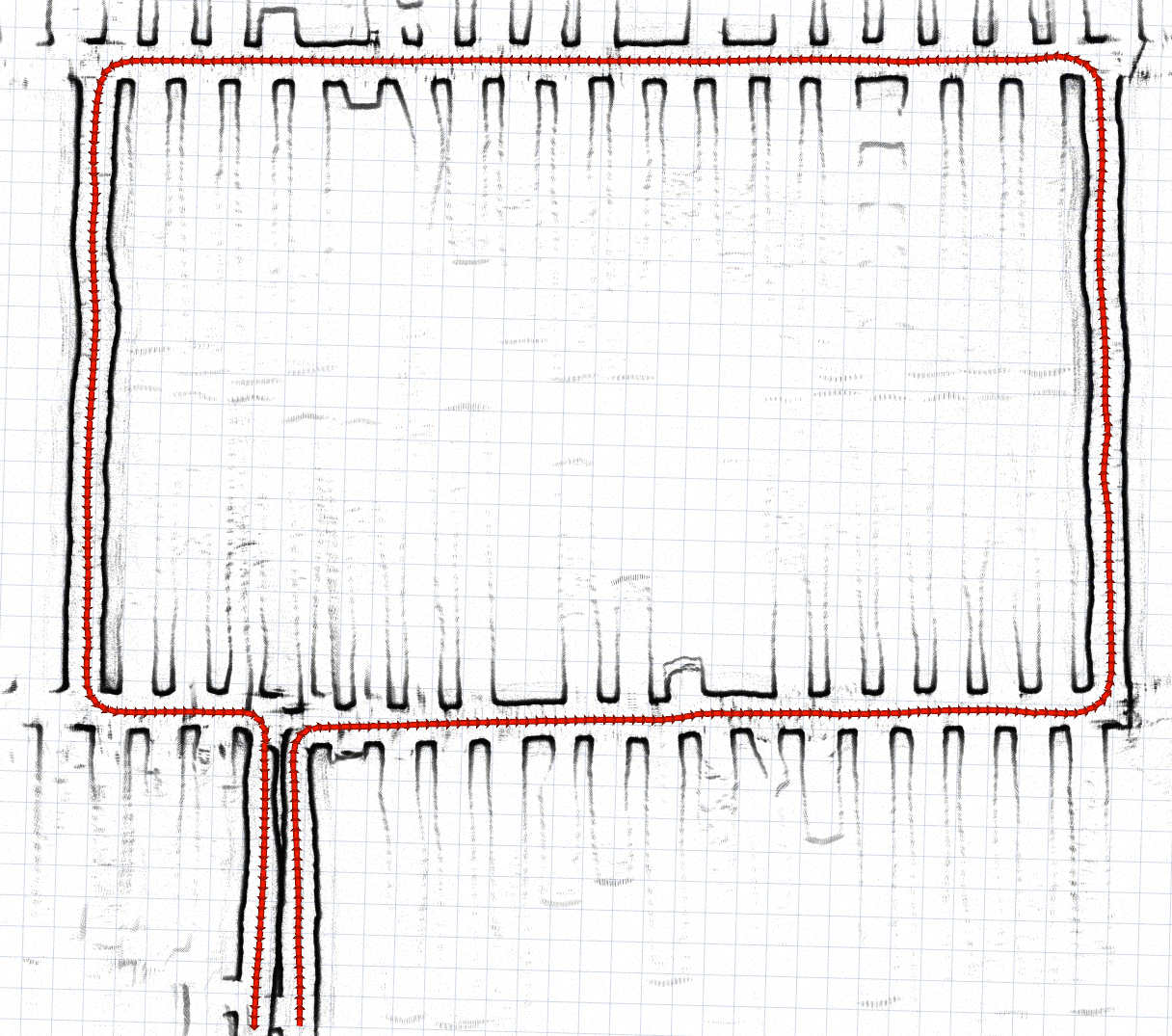}\label{fig:kvarntorp_odometry}}\hfill\\
    \includegraphics[trim={0.0cm 0cm 0cm 0cm},clip,height=0.29\hsize,angle=0]{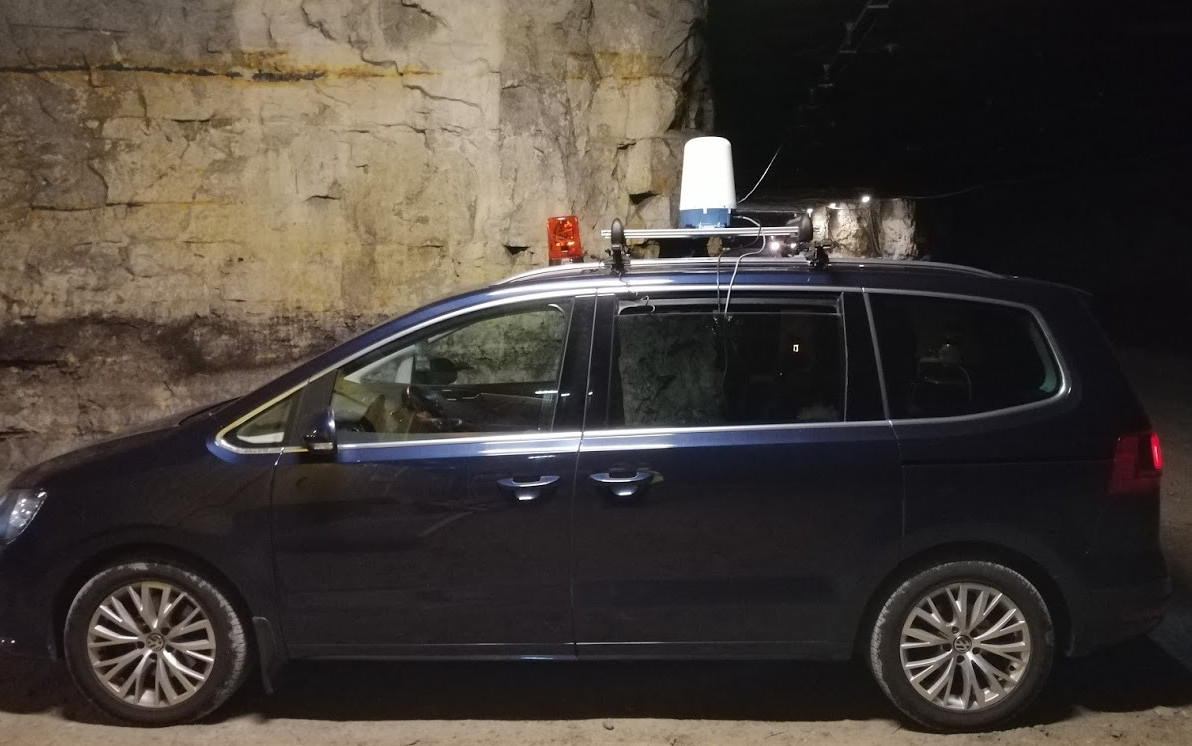}\label{fig:kvarntorp_vehicle}
      \includegraphics[trim={0.0cm 1cm 0cm 2cm},clip,height=0.29\hsize,angle=0]{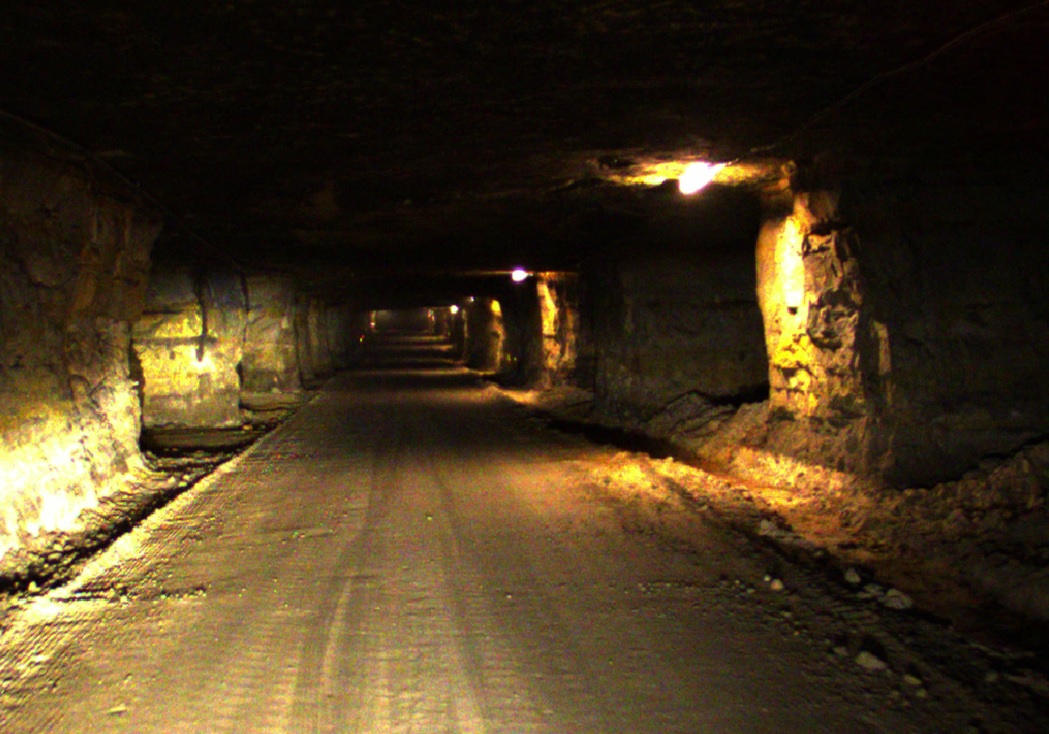}\label{fig:kvarntorp_mine}
    \caption{
    Example output using the proposed method. Incremental odometry estimated from a roof-mounted radar in an underground mine using the proposed method with no additional sensor. Parameters are tuned for a different sensor in an urban environment but generalize well. The accumulated error is roughly $15$m over a distance of $1235$m driving ($1.2$\%).
    Our system is demonstrated at: \url{https://youtu.be/np-ceNrAcNY}.
    }\label{fig:kvarntorp_example}
  \vspace{-0.3cm}
\end{figure}

State-of-the-art radar odometry systems put great effort into overcoming sensor-specific challenges such as speckle noise, ghost objects and false positives 
using deep learning methods to remove noise~\cite{barnes_masking_2020}, predict robust key points~\cite{barnes_under_2020}, and through robust data association algorithms~\cite{hong2020radarslam,8460687,8793990}. Scan matching is then based on sparse and efficient key-points~\cite{barnes_under_2020,hong2020radarslam,8460687,8793990} or dense radar images~\cite{barnes_masking_2020,9197231}.

 
While learning-based methods perform well on locations included in the training set, they provide limited generalization and, accordingly, their accuracy deteriorates at previously unseen locations~\cite{barnes_masking_2020} (despite being trained with data acquired from over 200~km of driving)~\cite{barnes_under_2020,barnes_masking_2020}.


 
We present \ac{TITLE}, a learning free method for efficient and accurate large-scale odometry estimation.
\ac{TITLE} filter radar data that creates a sparse representation in which the environment is modeled as a set of oriented surface points as depicted in Fig.~\ref{fig:k_strong_normals}. By keeping only the $k$ strongest power returns per measurement azimuth above the expected noise level, 
a conservative but sufficient set of observation points can be maintained while removing most radar noise. The filtered radar returns can be used to create a sparse set of surface points and normals that models the underlying geometry of important surface features. Scan registration is efficiently carried out by minimizing a point-to-line metric between correspondences, using Huber loss to achieve robustness to outliers.

The central contribution of this paper is an efficient, accurate and robust method tailored for \ac{FMCW} radar in large-scale environments, reaching state-of-the-art accuracy on a public benchmark data-set. Our results show that \ac{TITLE} method performs well in widely different environments (urban, forest, tunnel/mine) and can be used with different sensor models without changing a single parameter. It requires no training and runs on a single-core laptop CPU at $55$~Hz.





\section{Related work}
\label{sec:rw-lidar}\label{sec:rw-vision}

\subsection{Filtering radar data}
\label{sec:rw-filter}

Raw radar scans are often converted to point clouds using filtering techniques that distinguish between targets and background noise. Due to the challenging noise characteristics of radar data, various approaches have been investigated to remove noise. The traditionally most widely used filtering technique, \ac{CFAR}~\cite{vivet2013localization}, cannot be applied directly, as it assumes to know the noise distribution \textit{a priori}, which is not the case with \ac{FMCW} radar, especially when used in different environments. 
Moreover, \ac{CFAR} requires manual tuning of multiple parameters \cite{8793990,8460687}.

Another method is to simply set a fixed noise/signal threshold~\cite{mielle-2019-comparative}. However, doing so comes with the risk to remove important information since the noise background level varies greatly in different settings. 

Deep learning methods \cite{barnes_masking_2020,weston2019probably,8794014} have also been proposed for filtering but they require offline training and labeled data-sets as in \cite{barnes_masking_2020}. Wetson et al. \cite{weston2019probably} used lidar data for labeling applying a static world assumption. Aldera et al. \cite{8794014} used sensor coherence and wheel odometry for training.
In Marck et al.~\cite{marck2013indoor}, only the strongest power return per azimuth is kept.
In Fig.~\ref{fig:k_eval} we show that keeping only the strongest returns per azimuth makes the odometry estimation inaccurate. 

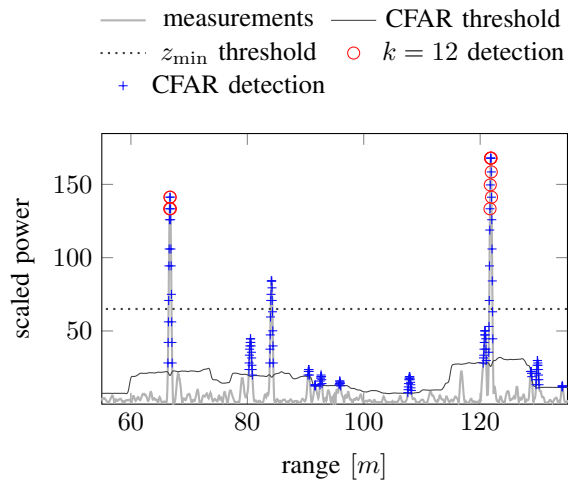
\begin{figure}
	\begin{tikzpicture}
	\begin{axis}
	[
		width   = 0.9\columnwidth,
		height  = 0.6\columnwidth,
		xmin		= 55,
		xmax		= 135,
		ymin		= 0.1,
		xlabel  = {range $[m]$},
		ylabel  = {scaled power},
		ylabel near ticks,
		legend style={
			at={(0.5,1.5)},
			anchor= north,
			fill = none,
			draw=none,
			},
		legend columns			= 2,
	]
	\addplot
	[
		white!70!black,
		thick,
	]
	table
	[
		x   = range,
		y   = beam,
	]
		{Datasets/export_figure_threshold.txt};
		\addlegendentry{measurements};
	\addplot
	[
		white!25!black,
		solid,
	]
	table
	[
		x   = range,
		y   = thr_cfar,
	]
		{Datasets/export_figure_threshold.txt};
		\addlegendentry{CFAR threshold};
	\addplot
	[
	white!25!black,
	thick,
	dotted,
	]
	table
	[
	x   = range,
	y   = thr_noise,
	]
	{Datasets/export_figure_threshold.txt};
	\addlegendentry{$\zmin$ threshold};
	\addplot
	[
	white!5!red,
	only marks,
	mark size	= 0.08cm,
	mark		= o,
	]
	table
	[
	x   = range,
	y   = k_12,
	]
	{Datasets/export_figure_threshold.txt};
	\addlegendentry{$k=12$ detection};
	\addplot
	[
	white!5!blue,
	only marks,
	mark size	= 0.06cm,
	mark		= +,
	]
	table
	[
	x   = range,
	y   = cfar,
	]
	{Datasets/export_figure_threshold.txt};
	\addlegendentry{CFAR detection};	
	\end{axis}
\end{tikzpicture}
	\caption{Typical power/range plot for a single azimuth reading. 
	Here we use $k=12$ for conservative filtering that reduces false positives at the cost of missing some landmarks by simply limiting the maximum number of detections per azimuth. Accordingly, $\zmin$ can be selected low make the filter sensitive without drastically increase the number of false detections. In contrast, 
	\ac{CFAR} returns more spurious detections and is hard to tune.\label{fig:k_threshold}}
	\vspace{-0.3cm}
\end{figure}

In this work, we extend the work of Marck et al.~\cite{marck2013indoor} by including only the strongest $k$ returns in a given beam that exceed the expected noise level $\zmin$.
Section~\ref{ssec:Parameters} presents an experimental evaluation of the effect of $k$ 
on the performance of radar odometry. The difference between the proposed CFEAR filtering and the classical \ac{CFAR} method can be seen in Fig.~\ref{fig:k_threshold}.


\subsection{Radar odometry}


\label{sec:rw-radar}
Over the years, various methods have been proposed for radar SLAM~\cite{6689216, Vivet_2013, 7795967}.  Recent radar odometry methods can be divided into sparse methods that attempt to estimate key points or compute corresponding features from radar images, and dense methods that match a denser set of radar returns. 
Holder et al. used ICP to align point clouds acquired from a front-facing radar~\cite{8813841} however the method relies on additional sensors.

Mielle et al.~\cite{mielle-2019-comparative} demonstrated radar odometry with off-the-shelf frameworks for range-based mapping in an indoor environment, using a fixed-threshold filtering method. 
However, the validation was done in a single, small-scale environment, and the fixed-threshold filtering has been found to be difficult to generalise to other environments. 

Cen et al. presented sparse methods for ego-motion~\cite{8460687,8793990}. A landmark extraction method detects peaks while removing multipath reflections. Data association and outlier rejection is carried out by finding the largest common subset of landmarks between subsequent scans. Landmarks are detected only in the polar space using 1d (power/range) signals. In contrast, our method additionally filters range data in Cartesian space and we obtain robustness to outliers via a Huber loss. Hong et al.~\cite{hong2020radarslam} proposed a full radar SLAM framework. The odometry method extract SURF feature descriptors without filtering the raw radar data and uses robust data association similarly to Cen~\cite{8460687}. 
Both Cen and Hong estimate the motion by minimizing the (weighted) squared error between correspondences using SVD (Single Value Decomposition), the performance of their methods is included in our evaluation.


Barnes et al. presented two learning based methods for radar odometry.
In~\cite{barnes_masking_2020} a mask is learned for filtering dynamic objects and noise and matching is carried out with a dense correlation-based approach. Barnes and Posner later proposed a sparse method that learns keypoints, scores and descriptors for odometry and loop closure detection~\cite{barnes_under_2020}. Keypoints are associated via a dense search and matched by minimizing the squared distance using SVD. Both methods~\cite{barnes_masking_2020,barnes_under_2020} require supervision from a previously existing positioning system and an extensive amount of data collected over $200$~km driving. Spatial cross-validation shows that the accuracy drops by a factor of $2.4$ when estimating odometry in new streets not included in the training set, yet in nearby locations within the same city~\cite{barnes_masking_2020}. It is currently not known how well~\cite{barnes_under_2020} generalizes to new locations. Our method does not rely on training or a previously existing positioning system. In the evaluation we compare these methods to ours and qualitatively demonstrate how our method generalizes to widely different environments.


Park et al.~\cite{9197231} used dense matching by applying the Fourier Mellin
transform (FMT) to Cartesian and log-polar radar images and estimating the relative motion from phase correlation in a course-to-fine fashion.
Their dense method yields poor run time performance and has not been compared to the state-of-the-art.

Radar odometry has also been improved by accounting for motion distortion~\cite{burnett_we_2021}, using multiple Doppler radars~\cite{6907064} and by introspectively detecting registration failures~\cite{8917111}. 

Finally, radar based localization with respect to a map have also been proposed in~\cite{narula_all-weather_2020,saftescu_kidnapped_2020,gadd2020look,8957240,tang2020selfsupervised}. However, these methods are out of scope as we focus on the radar odometry problem without a prior map.


\section{Rotating radar sensing}
\label{sec:method}
Our method \ac{TITLE} is tailored for, but not specific to, a rotating \ac{FMCW} radar without Doppler information. 
At the end of each full $360^\circ$ rotation, the sensor outputs a full power reading  $Z_{m \times n}$ in polar form, where $m$ is the amount of azimuths per sweep and $n$ is the amount of range bins per azimuth. 

\begin{figure}
\vspace{0.2cm}
    \centering
\includegraphics[width=0.99\hsize]{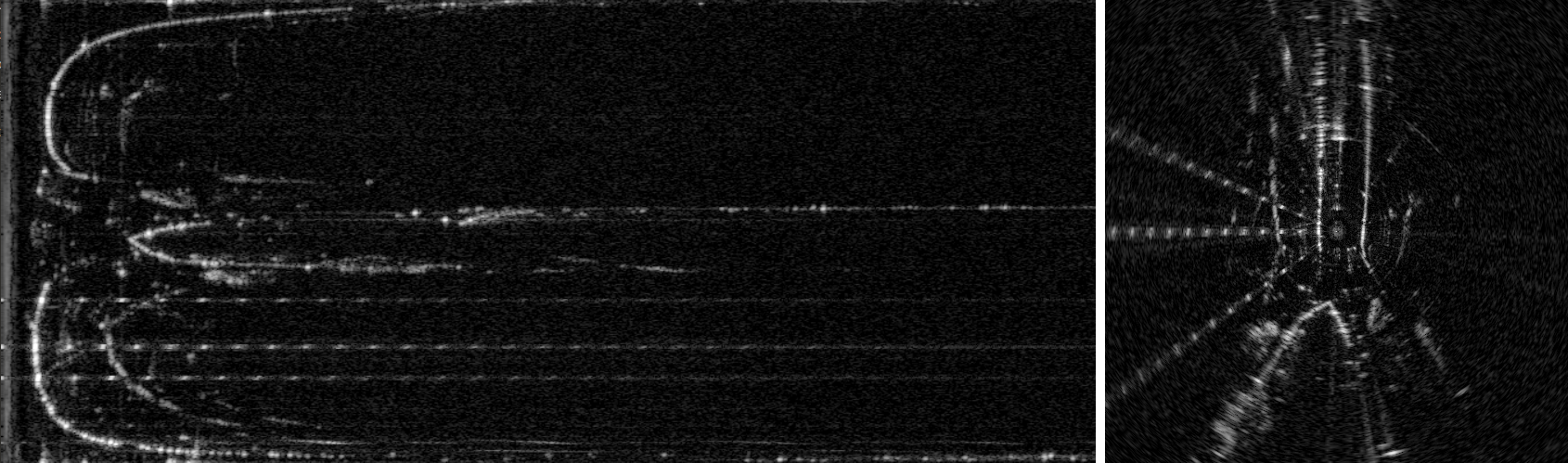}
    \caption{Conversion between polar (left) $\mathcal{Z}_{m\times n}$ and Cartesian (right) space. Each row and column in the polar image correspond to a range and angle $(a,d)$ in the Cartesian image.}
    \label{fig:polar_2_cart}
    \vspace{-0.2cm}
\end{figure}

Each azimuth and range bin $(a,d)$ in $Z_{m \times n}$ stores a power return and can be converted from polar to Cartesian space as seen in Fig.~\ref{fig:polar_2_cart} using Eq.~\eqref{eq:cartesian}.
\begin{equation}
\label{eq:cartesian}
    \mathbf{p} = \begin{bmatrix} d \gamma cos(\theta) \\ 
    d \gamma sin(\theta)
\end{bmatrix},
\end{equation}
where $\theta=2\pi a/m$ and $\gamma$ is the range resolution set to $4.38$cm or $17.5$cm in our experiments.

\section{\ac{TITLE}}
We argue that it is hard to find stable keypoints from peaks per azimuth due to large beam width, and radar sensing artifacts. 
In \ac{TITLE}, we compute a scan representation in two steps. First, only the $k$ strongest returns with power exceeding $\zmin$ are selected at each azimuth and converted into Cartesian space. Second, oriented surface points are computed in Cartesian space by analyzing local point distributions at discrete intervals. This creates a sparse surface representation from the originally noisy radar data, which rejects artifacts such as multi-path reflections in favor of important landmarks.

\subsection{k-strongest filtering}
The typical behavior of \textit{$k$-strongest filtering} compared to the classical method CFAR can be seen in Fig.~\ref{fig:k_threshold}. Keeping only the $k$ strongest returns exceeding the power $\zmin$ reduces false positives at the expense of eventually missing landmarks with weaker returns. 
However as shown by our evaluation depicted in Fig.~\ref{fig:k_eval}, missing weaker landmarks has a negligible impact on localization accuracy. This indicates that a conservative filtering technique is preferable for efficient radar odometry estimation.

\subsection{Estimating oriented surface points}
\begin{figure}
\vspace{0.2cm}
\begin{centering}
    \subfloat[CFAR keeps an excessive amount of points (black), including false positives, and surface points are poorly estimated.]
      {\includegraphics[trim={5cm 8.5cm 2cm 8cm},clip,width=0.9\hsize,angle=0]{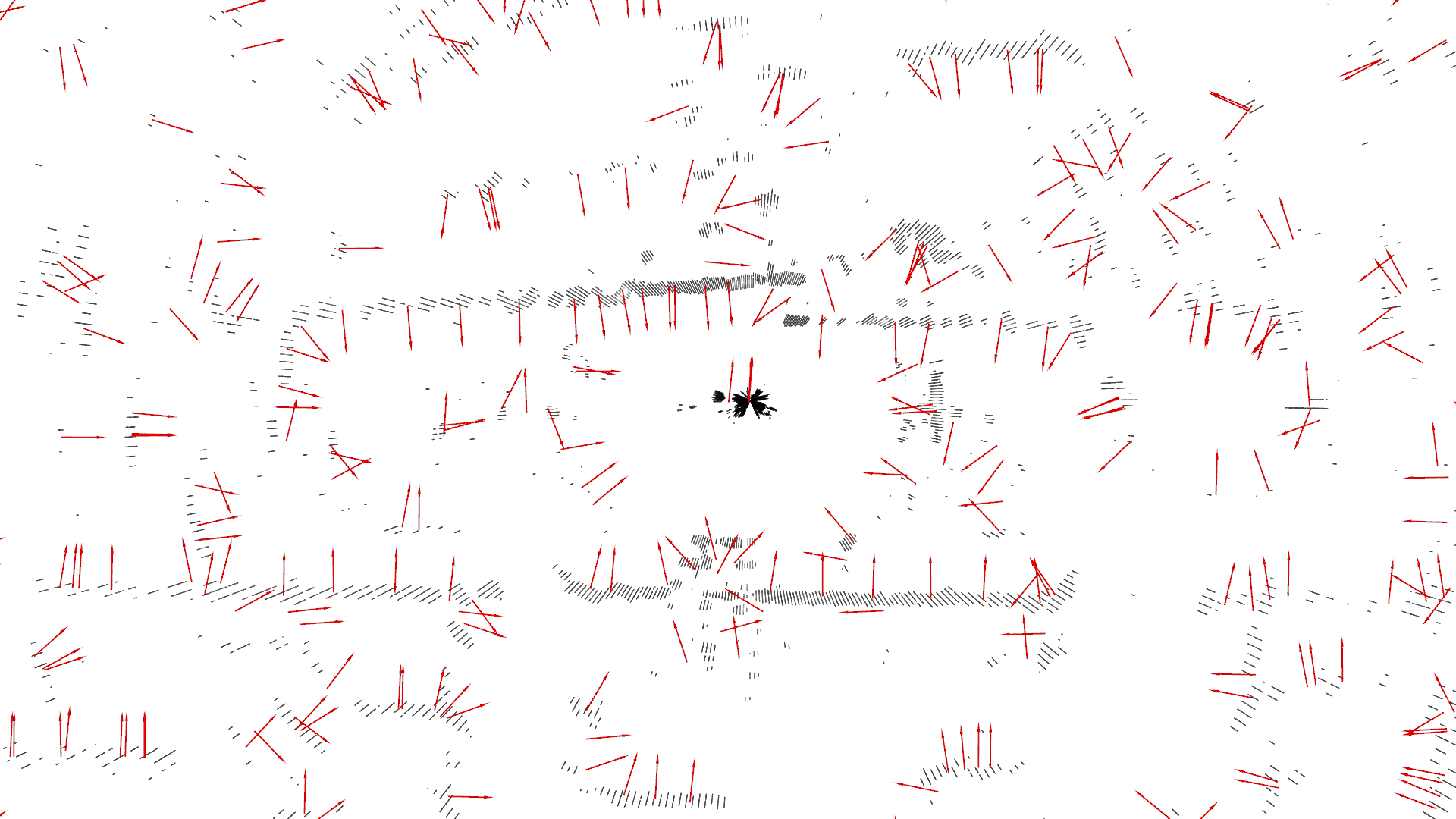}\label{fig:cfar_normals}}\\
    \subfloat[\ac{TITLE}: environment is represented by oriented surface points (red). The \textit{k-strongest} filtered points (fewer false positives) are shown in black.]
     {\includegraphics[trim={5cm 8.5cm 2cm 8cm},clip,width=0.99\hsize,angle=0]{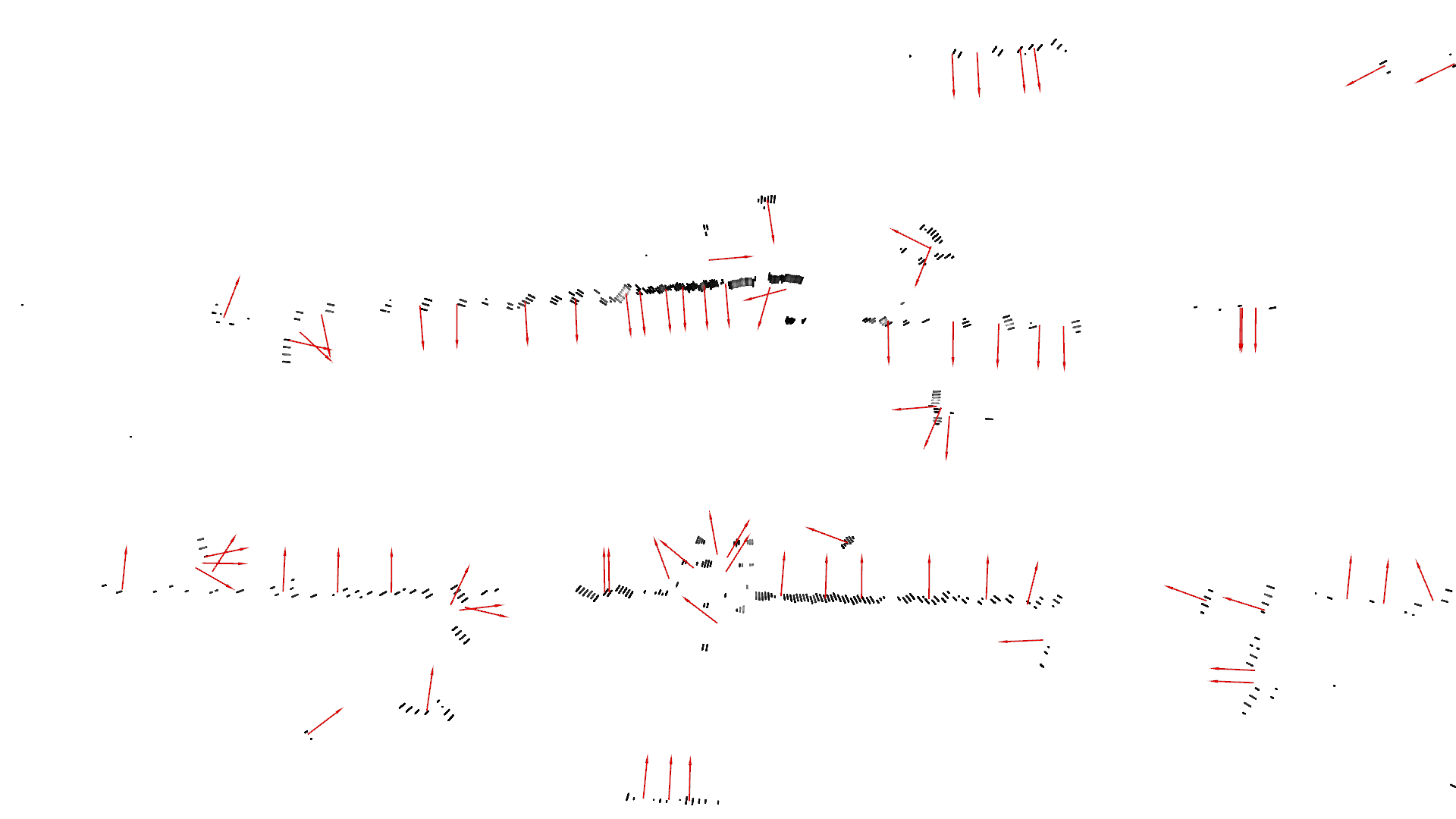}\label{fig:k_strong_normals}}
     \caption{Comparison of filtering methods, and oriented surface points.}
    \label{fig:comparsion_cfar_kstorng_normals}
  \end{centering}
\end{figure}

Given the filtered point set $\bm{\mathcal{P}}_f$
we aim to model the underlying surfaces in the environment by a set of oriented points; i.e., surface points and normals $\{\bm{\mu}_i,\mathbf{n}_i\}$.
This is done by first downsampling the point cloud $\bm{\mathcal{P}}_f$ to $\bm{\mathcal{P}}_d$ using a grid with side length $r/f$~m, with 
$r$ the distance in which we expect to find sufficient amount of points to estimate the normal, and $f$ a re-sampling factor that adjusts the density of the final representation.
For each grid cell 
only the centroid is kept.
For each point in the downsampled point cloud $p_i \in \bm{\mathcal{P}}_d$, all neighbors of $p_i$ in $\bm{\mathcal{P}}_f$ within a radius $r$ (same $r$ as above), are used to estimate the point distribution by the sample mean and covariance $\{\bm{\mu}_i,\mathbf{\Sigma}_i\}$. 
Ill-defined distributions where the condition number $\kappa(\mathbf{\Sigma}_i)=\lambda_\mathrm{max}(\mathbf{\Sigma}_i)/\lambda_\mathrm{min}(\mathbf{\Sigma}_i) > 10^5$, or where distributions are estimated from less than $6$ points are discarded. This works as an additional filtering step that removes remaining false detections after \textit{k-strongest} filtering and ensures that surface points are estimated from a sufficient number of points and from multiple azimuths. The surface normal $\mathbf{n}_i$ is obtained from the smallest of the two eigenvectors of $\mathbf{\Sigma}_i$. The full scan representation $\bm{\mathcal{M}}$ is obtained by pairing all surface points and normals $\bm{\mathcal{M}} = \{\mathbf{\bm{\mu}}_i,\mathbf{n}_i\}$. An example of reconstructed surfaces after applying the \textit{$k$-strongest filter} in comparison to CFAR can be seen in Fig.~\ref{fig:comparsion_cfar_kstorng_normals}.

\subsection{Multiple keyframe scan registration}
We employ a sliding window keyframe approach where the latest scan $\bm{\mathcal{M}}^t$ is registered to the $s$ most recent keyframes $\bm{\mathcal{M}}^{k_{1..s}}$. When the estimated relative pose 
exceeds a minimum rotation or translation to the latest keyframe, a new keyframe is stored from the current 
pose and point cloud $\{\bm{\mathcal{M}}^{k_s},\mathbf{T}^{k_s}\}\leftarrow\{\bm{\mathcal{M}}^t,\mathbf{T}^t\}$ and the oldest keyframe is removed. 
This ensures that drift is not introduced when the robot is stationary. Simultaneously registering to multiple keyframes can improve registration accuracy by imposing additional constraints, and improve robustness in cases when the registration is under-constrained or scan overlap drastically changes. Large changes in overlap can  occur due to occlusions from stationary or dynamic objects or due to non-planar road conditions~\cite{8917111}.

\begin{figure}
    \centering
    \includegraphics[width=0.25\textwidth]{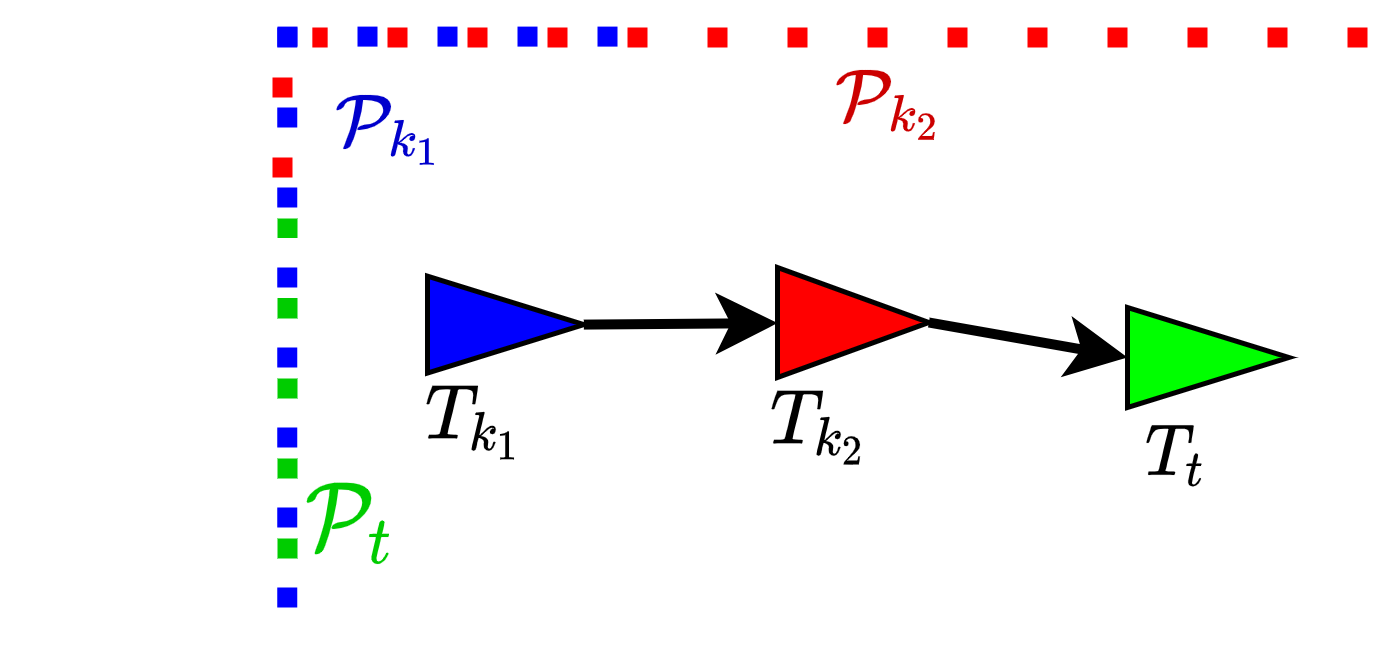}
    \caption{The latest scan $\bm{\mathcal{P}}^t$ is registered to the $s=2$ previous key-frames. This can improve accuracy due to noise and robustness due to occlusions. }
    \label{fig:n_keyframe}
\end{figure}

\begin{figure}
    \vspace{0.2cm}
    \centering
    \includegraphics[width=0.20\textwidth]{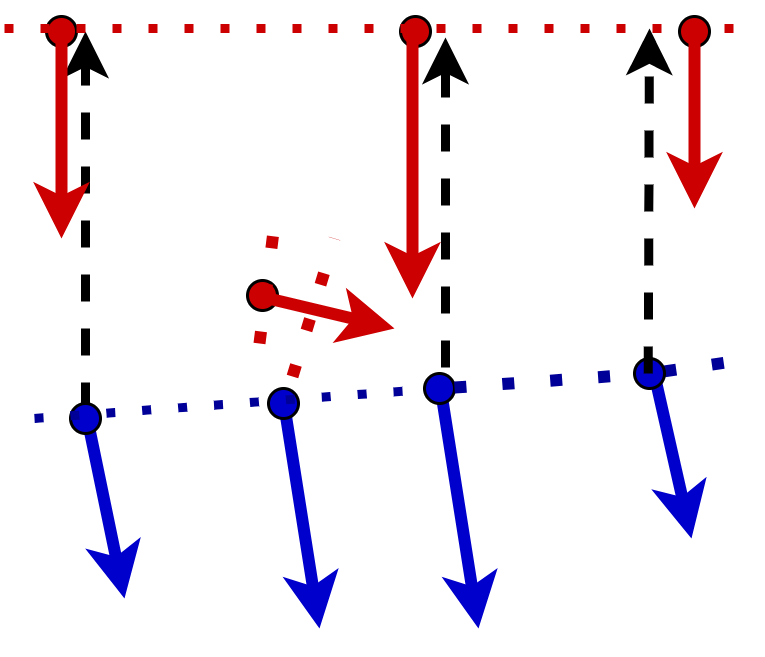}
    \caption{$\bm{\mathcal{M}}^t$ (blue) registered with $\bm{\mathcal{M}}^k$ (red). Most blue surface points (except one) have a correspondence with normals within $\theta_\mathrm{max}$ tolerance. The cost is given by the point-to-line distance visualized by black arrows.}
    \label{fig:point_to_line}
    \vspace{-0.2cm}
\end{figure}

We formulate the registration problem by minimizing a \textit{scan-to-keyframes (s2ks)} cost function 
\begin{equation}
    \label{eq:reg}
    \underset{\mathbf{x}}{\mathrm{arg\: min}} f_{s2ks}(\bm{\mathcal{M}}^{k_{1..s}},\bm{\mathcal{M}}^t,\mathbf{x}),
\end{equation}
where we seek the transformation parameters
$\bm{x}= \begin{pmatrix} x & y & \theta\end{pmatrix}$ 
that best align the current scan $\bm{\mathcal{M}}^t$ with the key-frames $\bm{\mathcal{M}}^{k_{1..s}}$ jointly 

\begin{equation}
    \label{eq:reg_individual}
     f_{s2k}(\bm{\mathcal{M}}^{k_{1..s}},\bm{\mathcal{M}}^t,\mathbf{x}) = \sum_{n=1}^{s}  f_{p2l}(\bm{\mathcal{M}}^{k_{n}},\bm{\mathcal{M}}^t,\mathbf{x}).
\end{equation}
The cost of aligning the latest scan $\bm{\mathcal{M}}^t$ to a single key-frame $\bm{\mathcal{M}}^k$ is given by the \textit{point-to-line (p2l)} function $f_{p2l}$ in Eq.~\eqref{eq:scan_to_keyframe}.

\begin{equation}
    \label{eq:scan_to_keyframe}
    f_{p2l}(\bm{\mathcal{M}}^k,\bm{\mathcal{M}}^t,\mathbf{x}) = \sum_{\forall i\in\bm{\mathcal{M}}^t} \mathcal{L}_\delta(\mathbf{n}^k_j \cdot (\mathbf{R}_{\theta}\bm{\mu}^t_i+\mathbf{t}_{x,y}-\bm{\mu}^k_j)),
\end{equation}
where each surface point $i\in\bm{\mathcal{M}}^t$ is associated with the closest surface point $j\in\bm{\mathcal{M}}^k$ within a radius $r$ if the angle between the surface normals is within a tolerance $\theta_\mathrm{max}$. $\mathbf{R}_{\theta}$ and $\mathbf{t}_{x,y}$ is the rotation matrix and translation vector created from the optimization parameters. An intuition for the score is visualized in Fig.~\ref{fig:point_to_line}.

Using the Huber loss $\mathcal{L}_\delta$, defined in Eq.~\eqref{eq:huber}, makes the cost less sensitive to outliers~\cite{Huber1992}. This is done by piecewise reshaping the cost function to increase quadratically for small values and linearly for larger values. 
\begin{equation}
    \label{eq:huber}
        \mathcal{L}_\delta(s)= 
\begin{cases}
    \frac{1}{2}s^2, & \text{if } s\leq \delta\\
    \delta(|s|-\frac{1}{2}\delta),              & \text{otherwise.}
\end{cases}
\end{equation}

We solve~\eqref{eq:reg} using the Broyden-Fletcher-Goldfarb-Shanno (BFGS) line search method with gradients obtained through automatic differentiation in the Ceres non-linear optimization framework~\cite{ceres-solver}.

\subsection{Accounting for motion}

At each frame the previous pose $\mathbf{T}^{t-1}$ and velocity estimate was used to predict a starting point $\mathbf{x}_\mathrm{init}$ for the optimization by linear extrapolation. The velocity was also used to compensate point cloud $\bm{\mathcal{P}}_f$ from motion distortion.
This is done by transforming all points into the time half way through the sweep, assuming that velocity is constant.


\section{Evaluation}\label{sec:evaluation}
In this section we investigate the performance of our method quantitatively on the public Oxford Radar RobotCar Dataset~\cite{RadarRobotCarDatasetICRA2020}, in an urban environment, 
and qualitatively on two data sets that we collected in non-urban environments (\textit{Kvarntorp mine} and \textit{Volvo test track}), to demonstrate how the method generalises without site and sensor specific parameter tuning. All experiments were carried out on an i7-7700HQ 2.80~GHz laptop CPU, running in a single thread.


\subsection{Parameters}
\label{ssec:Parameters}
The full set of parameters can be seen in Table~\ref{tab:Parameter}. We tuned the parameters to an urban environment using the sequence \textit{16-13-42} in the Oxford dataset. This sequence is not included in the evaluation set, with the intention to avoid fitting parameters to the specific traffic conditions in one of the evaluated sequences. After the parameter tuning, the exact same parameters were used through all experiments, despite widely different environments and although different radar models with different range resolutions were used.

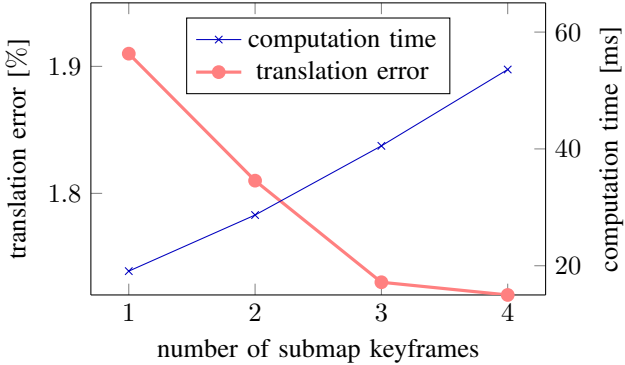
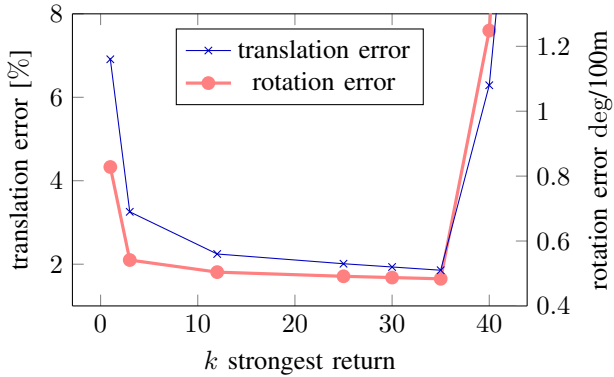
\begin{figure}
 \vspace{0.3cm}
  \begin{center}
    \subfloat[
    Increasing the amount of key-frames $s$
    reduces the translation error but requires additional processing time. Resample factor is set to $f=3$. ]{\begin{tikzpicture}

\pgfplotsset{width=7.5cm,compat=1.12}
  \pgfplotsset{
      scale only axis,
  }

  \begin{axis}[
	width   = 0.7\columnwidth,
	height  = 0.45\columnwidth,  
	ymin		= 1.72,
	ymax		= 1.95,
	legend style={at={(0.5,0.95)},anchor= north},
    axis y line*=left,
    xlabel={number of submap keyframes},
    ylabel={translation error [$\%$]},
  ]
    \addplot[mark=*,white!50!red,very thick]
      coordinates{
    (1.0000, 1.9100)
    (2.0000, 1.8100)
    (3.0000, 1.7300)
    (4.0000, 1.7200)
      }; \label{plot_1_y1}

    \end{axis}

    \begin{axis}[
		width   = 0.7\columnwidth,
		height  = 0.45\columnwidth,
	ymin		= 15,
	ymax		= 65,
	legend style={at={(0.5,0.95)},anchor= north},
    axis y line*=right,
    axis x line=none,
    ylabel={computation time [ms]},
    ]

    \addplot[mark=x,blue!75!black,thin]
      coordinates{
    (1, 19.0795)
    (2, 28.6807)
    (3, 40.5221)
    (4, 53.5934)
      }; \label{plot_1_y2}
    
    \addlegendimage{/pgfplots/refstyle=plot_1_y1}\addlegendentry{computation time}
    \addlegendimage{/pgfplots/refstyle=plot_1_y2}\addlegendentry{translation error}
  \end{axis}

\end{tikzpicture}\label{fig:keyframe_eval}}\hfill
    \subfloat[Within a wide range of values of \textit{k}, \ac{TITLE}  is insensitive to the exact value of \textit{k}.
        A too small or too large value, however, reduces translation and rotation accuracy. ]{\begin{tikzpicture}
\pgfplotsset{width=7.5cm,compat=1.12}
  \pgfplotsset{
      scale only axis,
  }

  \begin{axis}[
  	width   = 0.7\columnwidth,
	height  = 0.45\columnwidth,
	ymin		= 1,
	ymax		= 8,
	legend style={at={(0.5,0.95)},anchor= north},
    axis y line*=left,
    xlabel={$k$ strongest return},
    ylabel={translation error [$\%$]},
  ]
    \addplot[mark=*,white!50!red,very thick]
      coordinates{
    (1.0000,    4.3300)
    (3.0000,    2.1000)
    (12.0000,    1.8100)
    (25.0000,   1.7100)
    (30.0000,    1.6800)
    (35.0000,   1.6500)
    (40.0000,    7.6000)
    (50.0000,   41.6100)
      }; \label{plot_1_y1}

    \end{axis}

    \begin{axis}[
	width   = 0.7\columnwidth,
	height  = 0.45\columnwidth,
	ymin		= 0.4,
	ymax		= 1.3,
	legend style={at={(0.5,0.95)},anchor= north},
    axis y line*=right,
    axis x line=none,
    ylabel={rotation error $\deg$/100m},
    ]

    \addplot[mark=x,blue!75!black,thin]
      coordinates{
    (1.0000,  1.1600)
    (3.0000,  0.6900)
    (12.0000, 0.5600)
    (25.0000, 0.5300)
    (30.0000, 0.5200)
    (35.0000, 0.5100)
    (40.0000, 1.0800)
    (50.0000, 4.1900)
        
      }; \label{plot_1_y2}
    
    \addlegendimage{/pgfplots/refstyle=plot_1_y1}\addlegendentry{translation error}
    \addlegendimage{/pgfplots/refstyle=plot_1_y2}\addlegendentry{rotation error}
  \end{axis}

\end{tikzpicture}
    \label{fig:k_eval}}\hfill
    \caption{Accuracy/computation time using different parameters.}\label{fig:parameter_eval}
  \end{center}
  \vspace{-0.3cm}
\end{figure}

A brief evaluation of the most important parameters (performed on the Oxford dataset) can be seen in Fig.~\ref{fig:parameter_eval}. 
The results in Fig.~\ref{fig:keyframe_eval} show that the accuracy can be improved by increasing $s$ (submap keyframes) at the expense of runtime performance. 
Fig.~\ref{fig:k_eval} shows the effect of varying the number $k$ of returns in our filtering method.
Using too few ($k<12$) or too many ($k>35$) points negatively impact accuracy, but within the wide range $12\leq k \leq 35$ the method is largely insensitive to $k$.

\begin{table}
\centering
\begin{tabular}{l|lll}
\\
Parameters & Value          \\    
\hline
$k$                              & $12$ \\ 
Resample factor $f$                     & $1$ \\  
$\zmin$                               & $55$ \\  
Resolution $r$                          & $3.5$ \\  
$\theta_\mathrm{max}$                          & $30$ \\    
Huber loss    $\mathcal{L}_\delta$      & $0.1$ \\    
Key-frame (dist[m]/rot[deg])            & $1.5/5^\circ$ \\  
$s$                        & $3$ \\   
min. sensor distance [m]                 & $5$ \\     
max. sensor distance [m]                & $100$ \\   
\end{tabular}
\caption{Full list of parameters. These are used for all datasets and sensors, except when explicitly mentioned in Sec.~\ref{sec:volvo_eval}. }\label{tab:Parameter}
\end{table}

\subsection{Quantitative evaluation in urban environment}
\label{sec:quantitative_eval_sec}
We provide a quantitative evaluation of the odometry accuracy and performance in a large-scale environment with ground truth positioning. The odometry accuracy was measured as proposed in the KITTI odometry benchmark~\cite{Geiger2012CVPR} using the tool~\cite{zhan2019dfvo} that computes the average translation error ($\%$) and rotation error (deg/m) over all sub sequences between $\{100,200, \ldots, 800\}$~m.

We used the public Oxford Radar RobotCar Dataset which was collected with a roof-mounted Navtech CTS350-X \ac{FMCW} radar running at $4$~Hz, configured with a range resolution of $\gamma=4.38$~cm. The dataset contains 32 traversals through a $10$-km urban environment in varied weather, lighting and traffic conditions.

\begin{figure*}[]
  \begin{center}
    \subfloat[10-12-32]{\includegraphics[trim={0.0cm 0cm 0cm 0cm},clip,width=0.32\hsize]{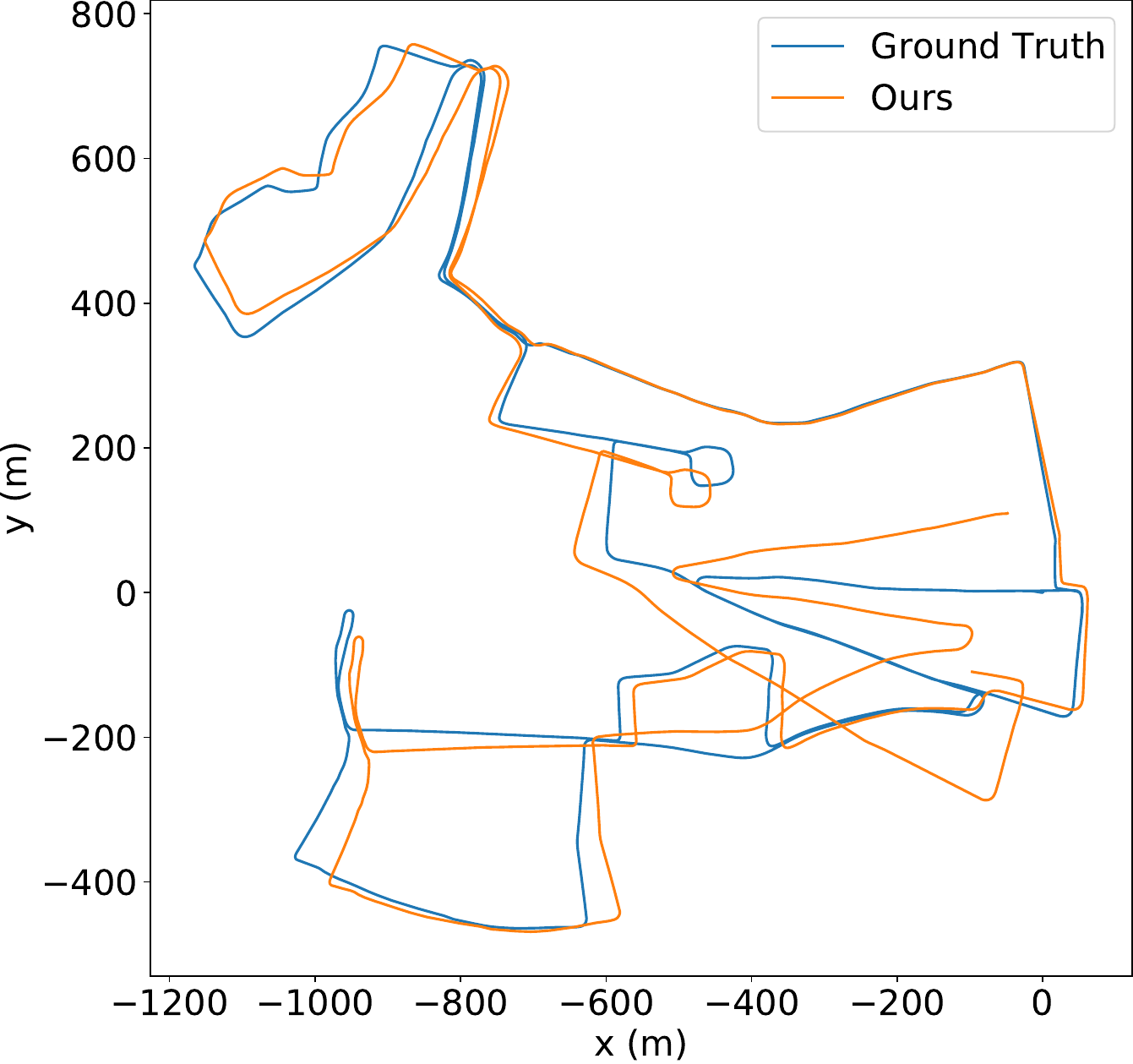}\label{fig:10-12-32}}\hfill
    \subfloat[16-13-09]{\includegraphics[trim={0.0cm 0cm 0cm 0cm},clip,width=0.32\hsize]{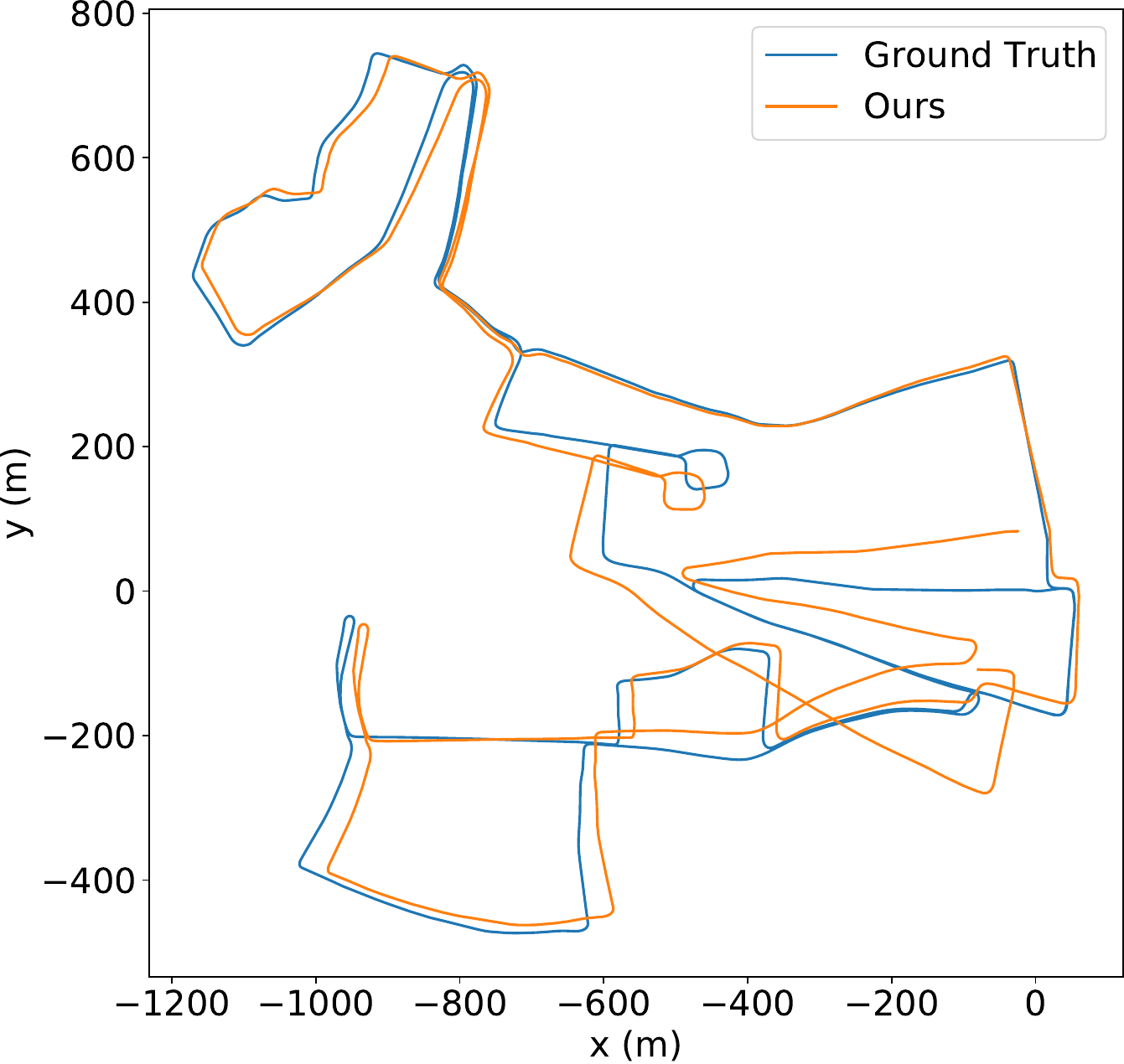}\label{fig:16-13-09}}\hfill
    \subfloat[17-13-26]{\includegraphics[trim={0.0cm 0cm 0cm 0cm},clip,width=0.32\hsize]{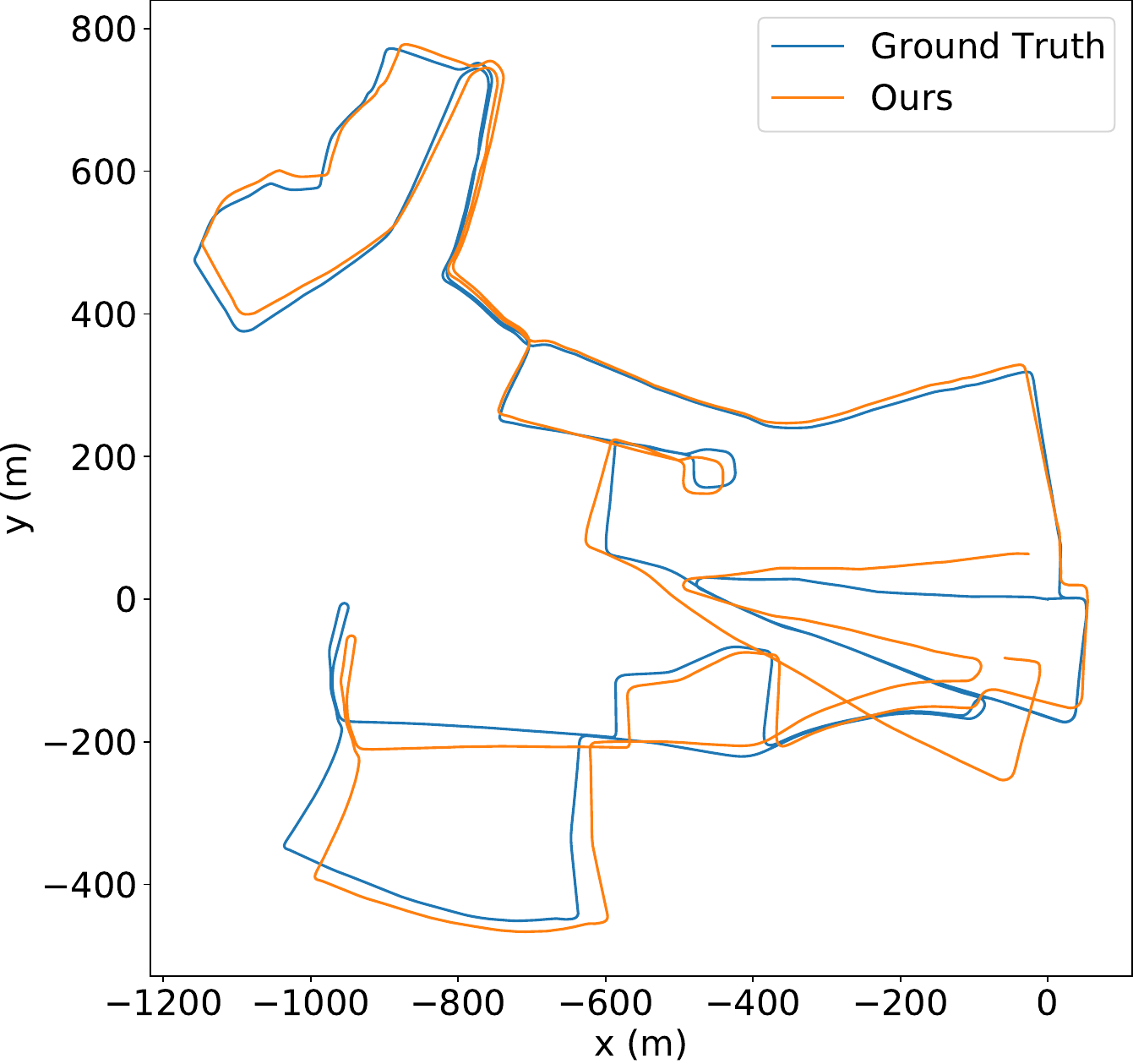}\label{fig:17-13-26}}\\\
    \vspace{-0.2cm}
     \subfloat[18-14-14]{\includegraphics[trim={0.0cm 0cm 0cm 0cm},clip,width=0.32\hsize]{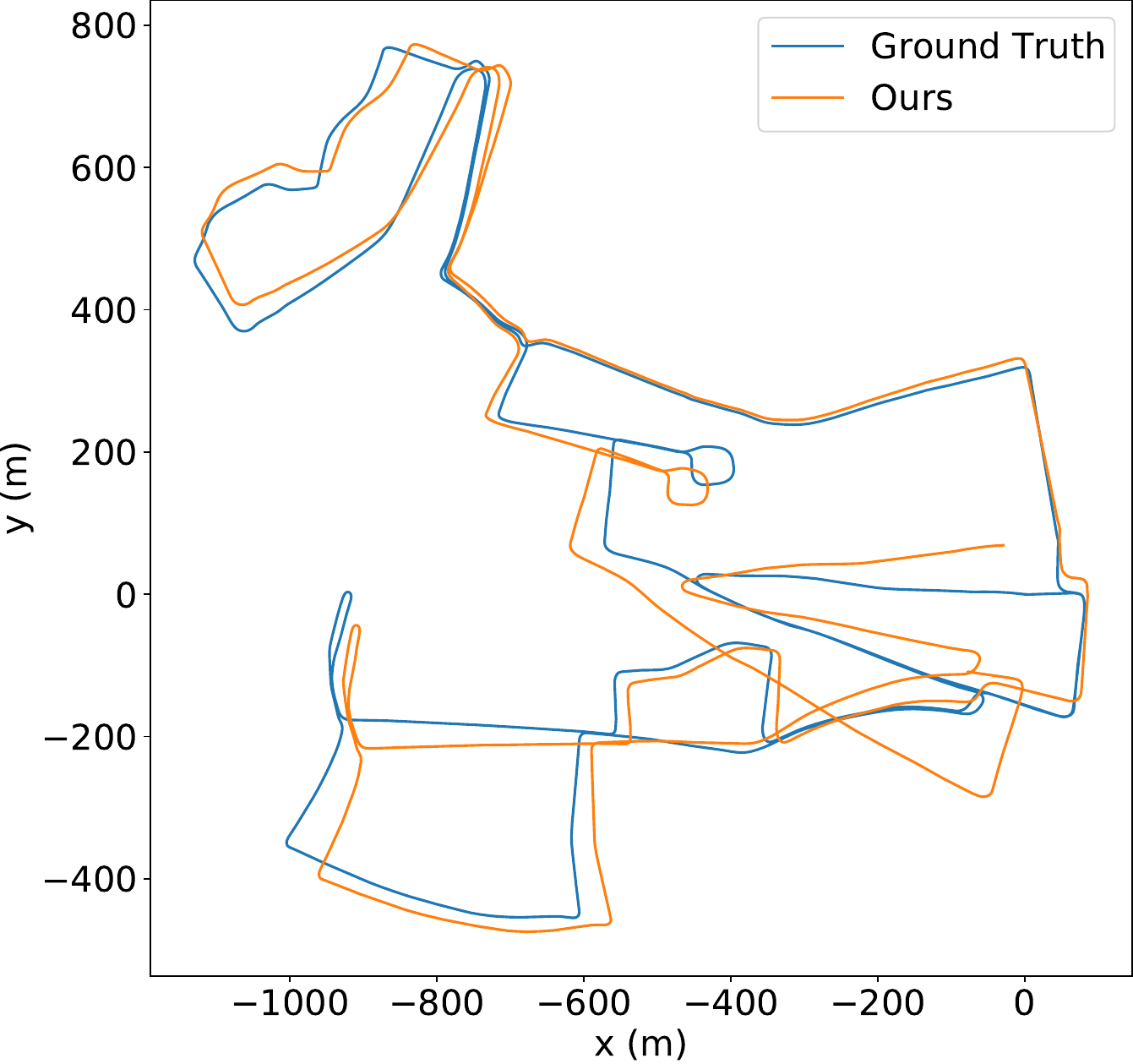}\label{fig:18-14-14}}\hfill
    \subfloat[18-15-20]{\includegraphics[trim={0.0cm 0cm 0cm 0cm},clip,width=0.32\hsize]{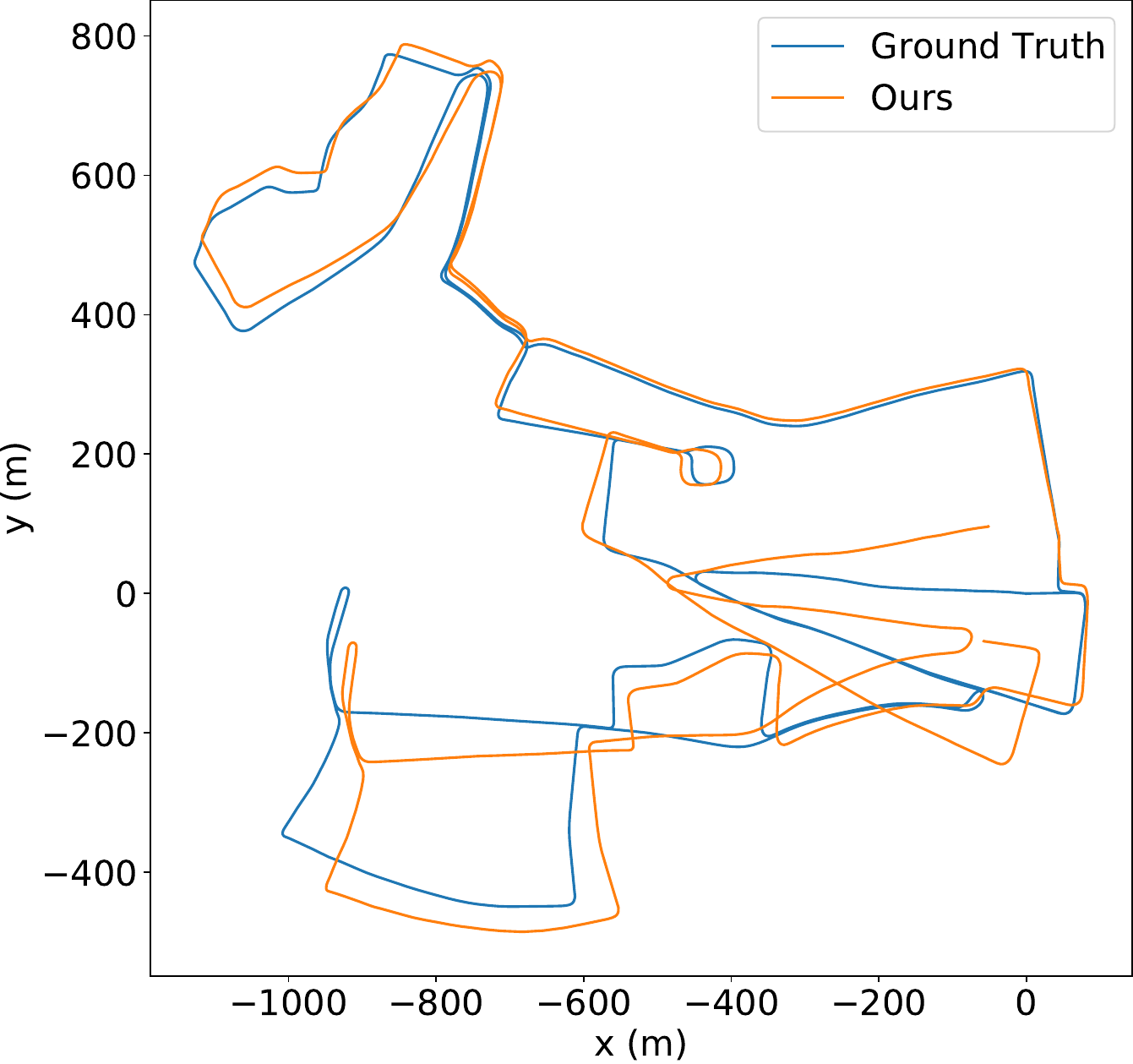}\label{fig:18-15-20}}\hfill
    \subfloat[16-11-53]{\includegraphics[trim={0.0cm 0cm 0cm 0cm},clip,width=0.32\hsize]{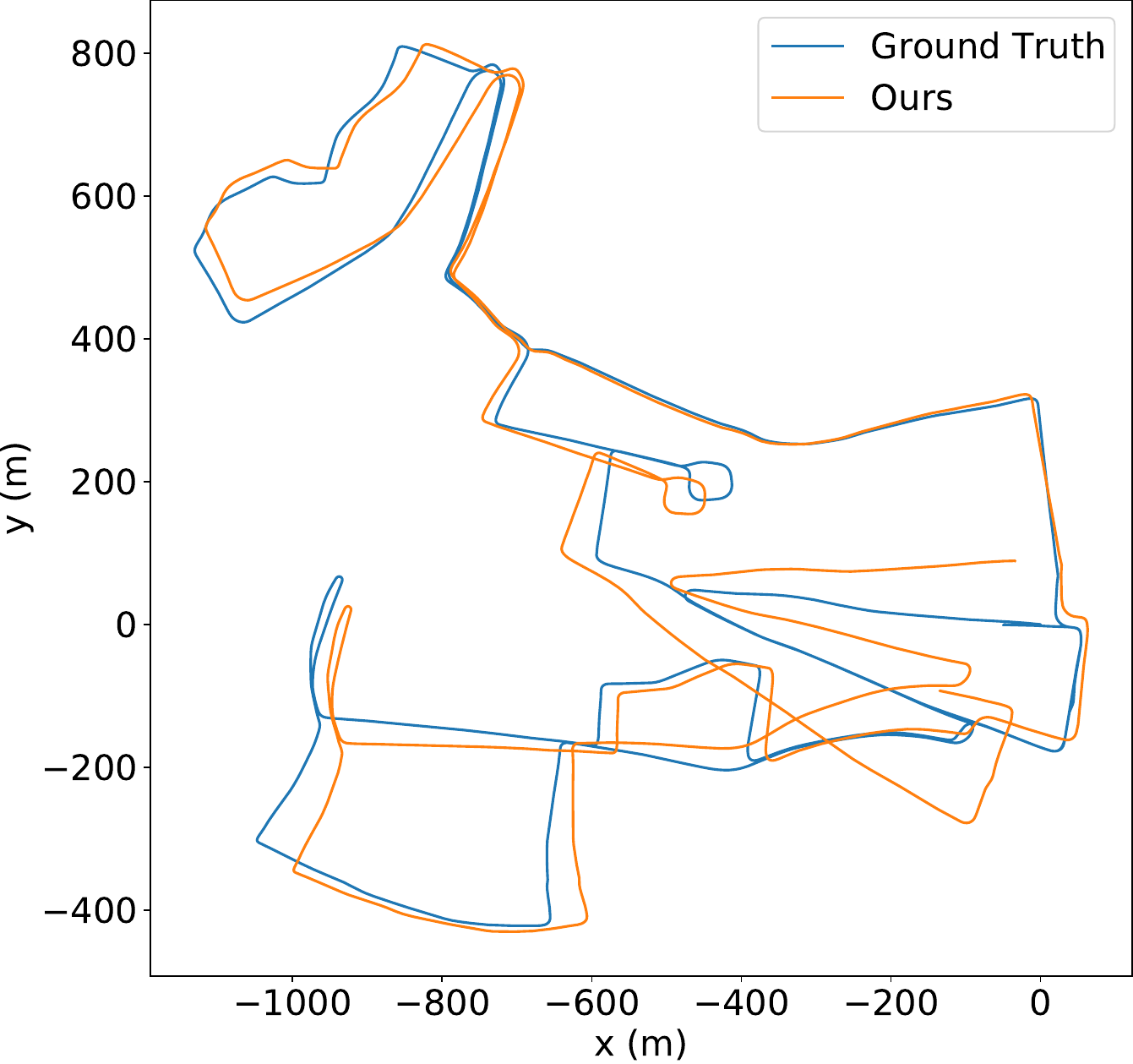}\label{fig:16-11-53}}
    \caption{Evaluated Oxford sequences using \ac{TITLE}. For qualitative comparison to other methods we refer to ~\cite{hong2020radarslam}.}\label{fig:sequences_oxford}
  \end{center}
  \vspace{-0.3cm}
\end{figure*}

In order to compare our results, we selected the same sequences that were used in the evaluation by Hong~\cite{hong2020radarslam}, both for quantitative evaluation as presented in Tab.~\ref{tab:results} and the visualization in Fig.~\ref{fig:sequences_oxford}. Learning-based methods are compared using their SCV (Spatial Cross Validation). With SCV, test and training datasets are taken from different parts of the environment, rather than selecting the training data from different traverses in the same locations. Without SCV, test and training datasets are not independent since test and training points may be neighbors~\cite{lovelace_geocomputation_2019}.


As can be seen in Tab.~\ref{tab:results}, \ac{TITLE} produces the lowest per-sequence and mean SCV error with a translation and rotation error of $1.76\%$ and $0.5$~deg/100m, followed by Barnes Dual Cart~\cite{barnes_masking_2020} ($2.78$\%), Hong odometry~\cite{hong2020radarslam} ($3.11$\%), and Cen ~\cite{8460687} ($3.72$\%).
\textit{Barnes Dual Cart}~\cite{barnes_masking_2020} achieves greater mean accuracy ($1.16$\% error) but only when training and evaluating on the same spatial location. When the authors evaluated their methods on places not seen in the training set, the error increased to from $1.76\%$ to $2.78$\%.
\ac{TITLE} provides a $14$\% lower mean odometry error compared to ~\cite{barnes_under_2020} without SCV. The method does not report SCV error in the original publication and it is unclear how well the method generalizes to new nearby locations or environments. 

We therefore argue that \ac{TITLE} method improves state-of-the-art accuracy for sparse methods, but also improves state-of-the-art among all radar odometry methods conditioned on that the evaluation is not biased with spatially overlapping training and evaluation data.

 \begin{figure}
  \vspace{0.2cm}
  \begin{center}
\includegraphics[width=0.89\hsize,trim={0cm 0cm 0cm 0cm},clip,angle=270]{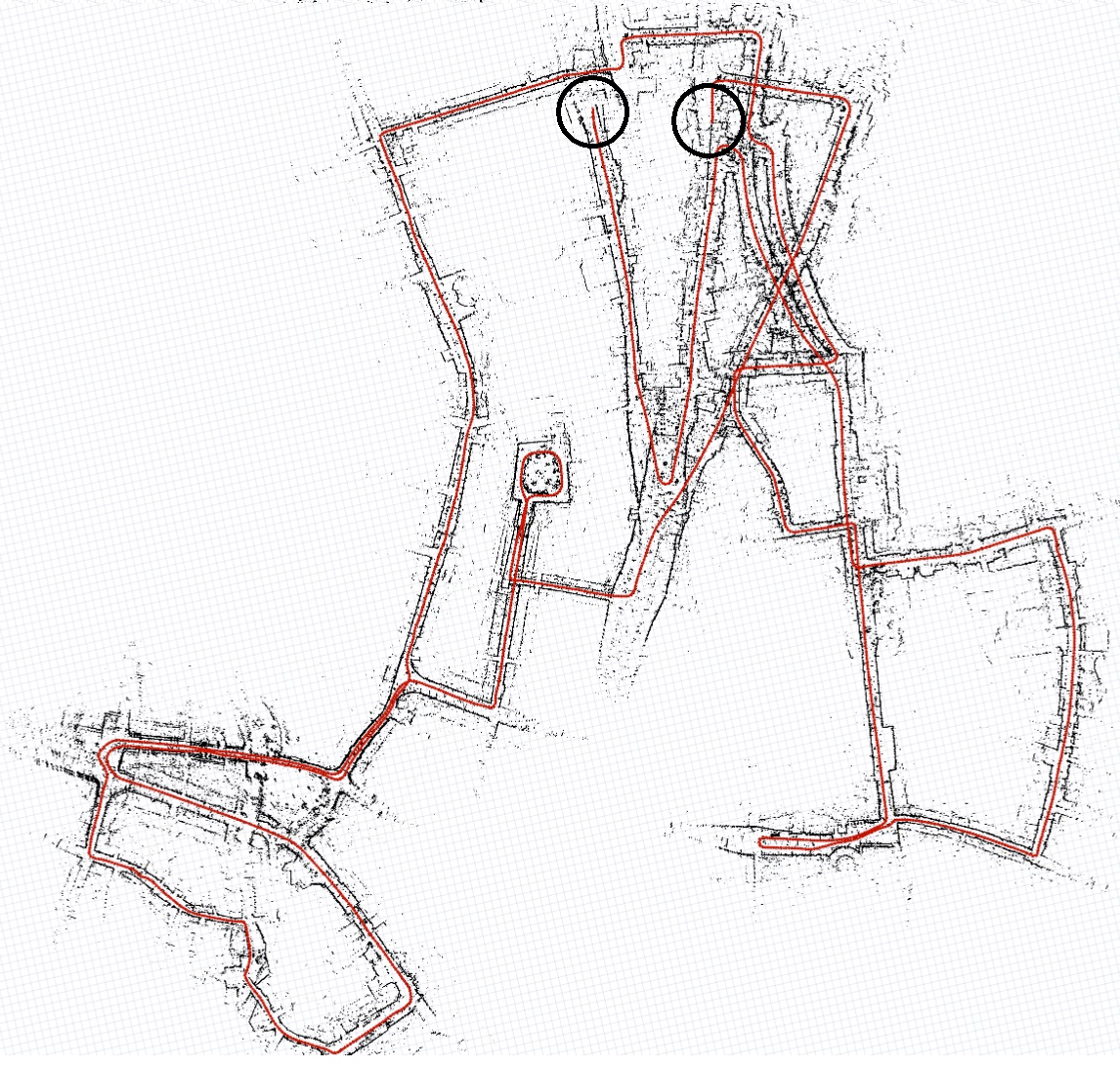}\caption{\label{fig:oxford_overview}Incremental radar odometry using the proposed method over a $10$~km sequence from the Oxford dataset. The first and final pose estimate is highlighted and gives an indication of the total aggregated drift. }
  \end{center}
 \vspace{-0.6cm}
\end{figure}


\begin{table*}
\centering
  \begin{adjustbox}{width=\textwidth}

\begin{tabular}{l|ll|lllllllll|l|l}
              & & & \multicolumn{1}{l}{\textbf{Sequence}}    &                                                                        \\
\textbf{Method} & \textbf{Evaluation} & \textbf{resolution}        & 10-12-32 & 16-13-09 & 17-13-26 & 18-14-14 & 18-15-20 & 10-11-46 & 16-11-53 & 18-14-46 & Mean  & \textbf{Mean SCV} & runtime (ms)   \\
\hline \\
Visual Odometry  ~\cite{Churchill2012ExperienceBN} & ~\cite{barnes_masking_2020} & N/A   & N/A        & N/A        & N/A        & N/A        & N/A        & N/A        & N/A        & N/A        &    $3.78/0.01$ & N/A   &      \\
\hline \\
SuMa (Lidar)~\cite{behley2018rss}  & ~\cite{hong2020radarslam }& N/A  & $1.1/0.3$ & $1.2$/$0.4$ & $1.1/0.3$ & $0.9/0.1$ & $1.0/0.2$  & $1.1/0.3$        & $0.9/0.3$        & $1.0/0.1$        & $1.16/0.3$    & $1.03/0.3$  &       \\
\hline \\
Cen~\cite{8460687}    &   ~\cite{barnes_masking_2020} & $0.175$& N/A        & N/A        & N/A        & N/A        & N/A        & N/A        & N/A        & N/A        & $3.72/0.95$    & $3.63/0.96$ &        \\
Robust Keypoints~\cite{barnes_under_2020}    &  & $0.346$    & N/A        & N/A        & N/A        & N/A        & N/A        & N/A        & N/A        & N/A        & $2.05/0.67^*$    & N/A  &         \\
Barnes Dual Cart~\cite{barnes_masking_2020} & & $0.043$ & N/A        & N/A        & N/A        & N/A        & N/A        & N/A        & N/A        & N/A        & $\bm{1.16/0.3}^*$    &   $2.784/0.85 $ &\\
Hong odometry~\cite{hong2020radarslam} & &$0.043$ & $2.98/0.8$        & $3.12/0.9$        & $2.92/0.8$        & $3.18/0.9$        & $2.85/0.9$        & $3.26/0.9$        & $3.28/0.9$        & $3.33/1$        & $3.11/0.9$    & $3.11/0.9$ &        \\
\ac{TITLE} (ours) &  & $0.043$ & \textbf{1.64/0.48}          & \textbf{1.86/0.52}        & \textbf{1.66/0.48}        & \textbf{1.71/0.49}        & \textbf{1.75/0.51}        & \textbf{1.65/0.48}        & \textbf{1.99/0.53}        & \textbf{1.79/0.5}        & 
1.76.0.50    & 
$\bm{1.76\pm0.12/0.50\pm0.02}$  & $18\pm 0.4$ 
\end{tabular}
  \end{adjustbox}

\caption{Evaluation on 8 sequences with different methods and sensor modalities on the Oxford Radar RobotCar dataset~\cite{RadarRobotCarDatasetICRA2020}.  Results are given in (\% translation error / deg/$100$~m). Note the difference between columns "Mean" and "Mean SCV". In "Mean" column, methods marked $^*$ cannot be compared directly as these are trained and evaluated on the same spatial location. The most relevant number for comparison is instead "Mean SCV" which ensures that test and training data are not correlated by using spatial cross validation. 
}
\label{tab:results}
\vspace{-0.2cm}
\end{table*}





\subsection{Qualitative evaluation}
In order to assess the generality of the method, additional non-urban  data was collected. We perform a qualitative evaluation of radar odometry in two different industrial scenarios using 
another Navtech \ac{FMCW} radar model,
CIR154XH. 
For these datasets, the sensor was configured with range resolution $\gamma=17.5$cm, with $400$ azimuth bins. Ground truth positioning is missing in these datasets. However, we estimate the error between the first and final scan by comparing the offset between duplicated landmarks in the map which gives a rough indication of odometry drift.

\subsubsection{Evaluation in semi-structured outdoor environment -- Volvo CE}
\label{sec:volvo_eval}

\begin{figure} 
 \vspace{0.2cm}
  \begin{center}
    \subfloat[Vehicle test track. A wheel loader with radar starts inside the white tent (right) and is driven in a large loop on a gravel road and then in two smaller loops on an uneven path in the forest. The full driven trajectory (seen in red) is imported from Fig.~\ref{fig:volvo_full_submap}.
    The locations of the photos in figure (b) and (c) are indicated along the trajectory.
    ]%
    {\begin{tikzpicture}
    \node[anchor=south west,inner sep=0] at (0,0) {\includegraphics[trim={0.0cm 0cm 0cm 0.2cm},clip,width=0.99\hsize]{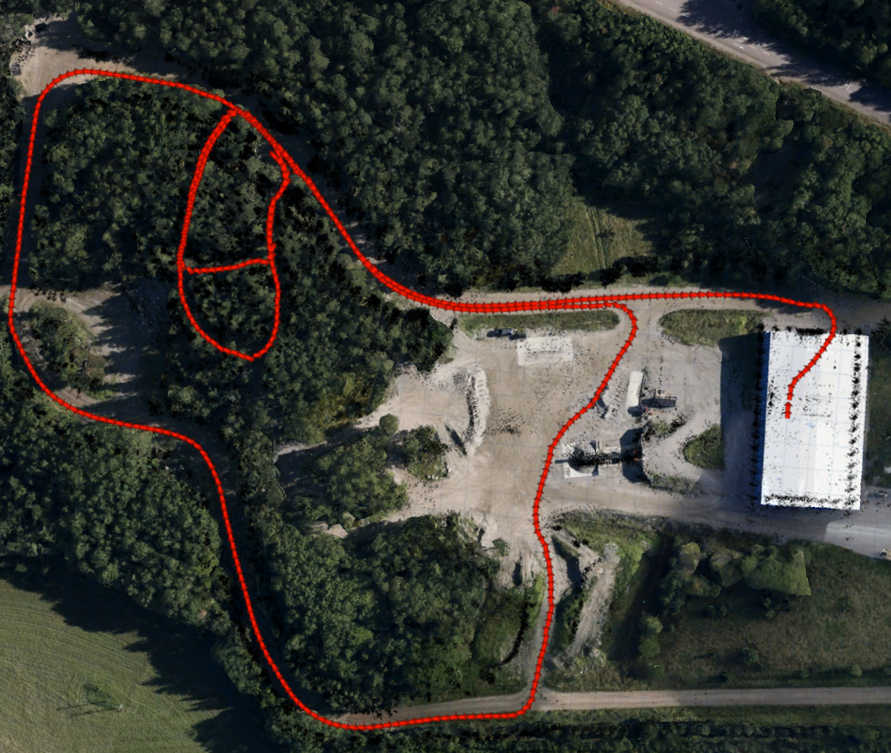}};
    \node[white!70!white] at(1.6,5.7) {(\textbf{b})};
    \node[white!70!white] at(3.5,5.3) {(\textbf{c})};
    \end{tikzpicture}}
    \\
        \subfloat[Driving on a narrow uneven path in the forest. ]{\includegraphics[trim={0.0cm 0cm 0cm 0cm},clip,height=43mm]{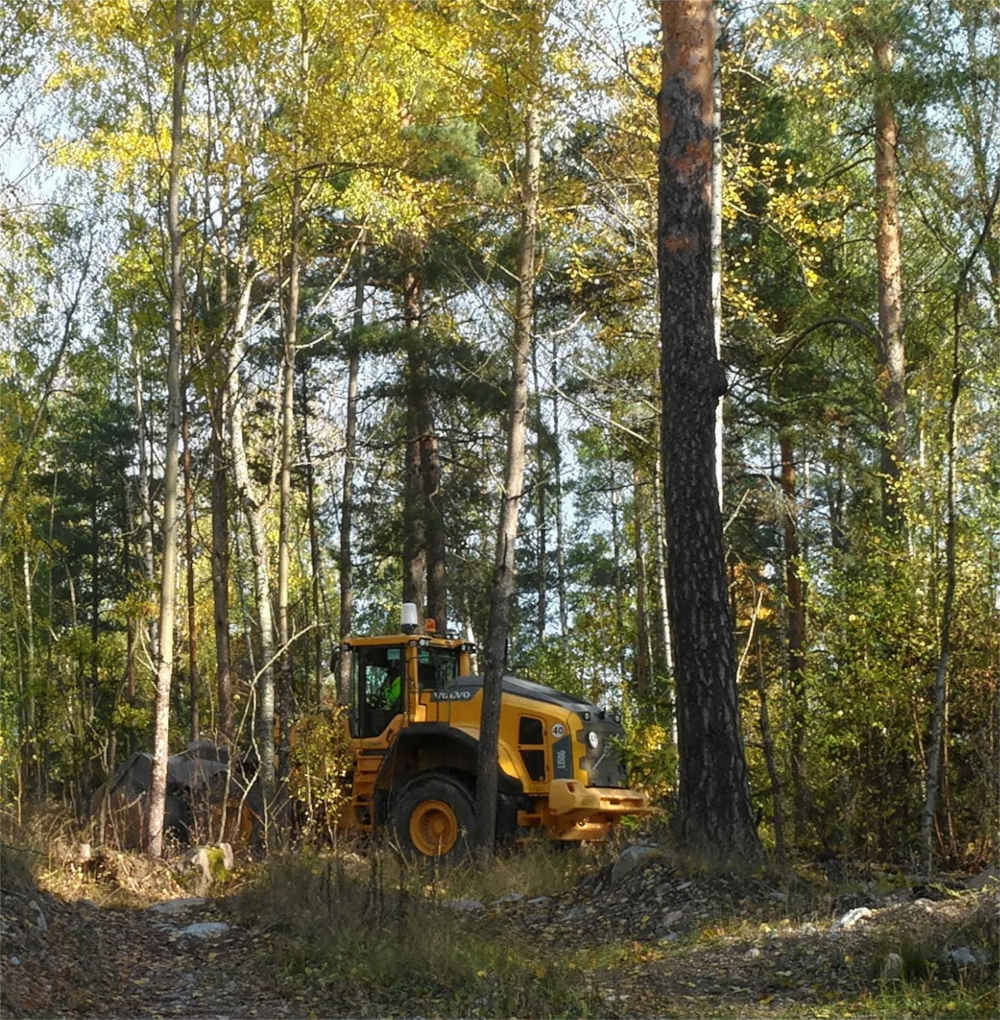}\label{fig:volvo_forest}}\hfill
        \subfloat[Driving on a gravel road uphill.]{\includegraphics[trim={0.0cm 0cm 0cm 0cm},clip,height=43mm]{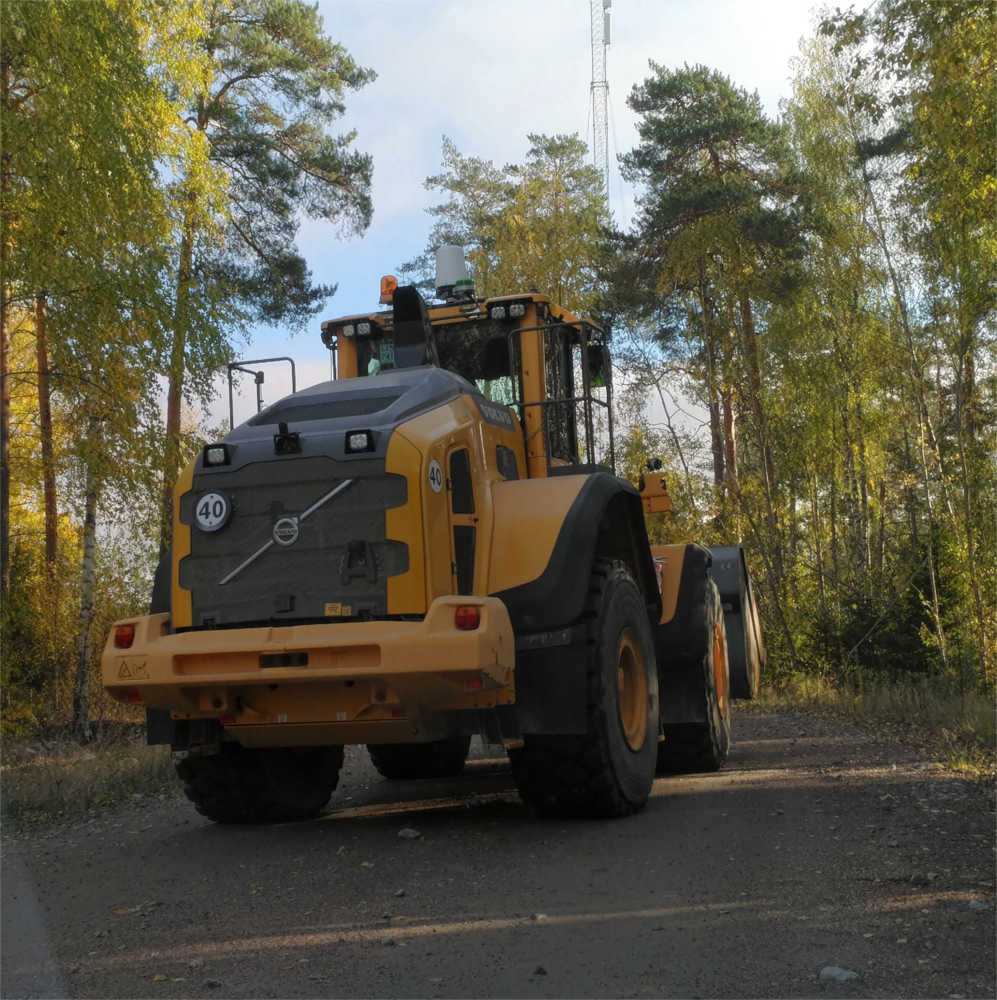}\label{fig:volvo_road}}\hfill \\
    
    \caption{Overview of semi-structured VolvoCE dataset.\label{fig:volvo_test_track_and_vehicle}}
  \end{center}
  \vspace{-0.8cm}
  \end{figure}

The first dataset was collected from a test track for wheel loaders and dump trucks in a partly forest environment driven at an approximate speed of $10$km/h. The track consists mainly of gravel roads at various slopes and the radar sensor was mounted on the driver cabin of a Volvo wheel loader (see Fig.~\ref{fig:volvo_test_track_and_vehicle}). The larger loop in Fig.~\ref{fig:volvo_test_track_and_vehicle} is $1150$ meters long. After completing the larger gravel-road loop, two smaller loops were traversed on uneven paths inside the forest reaching a total trajectory length of $1605$ meters. We expect this highly cluttered environment to be challenging for \ac{TITLE} that attempts to reconstruct and utilize surface normals.

\begin{figure}
\vspace{0.2cm}
  \begin{center}
    \subfloat[Full estimated trajectory (red) and map (black), obtained reusing parameters tuned for the Oxford urban dataset. The maximum estimated large loop error is $15$~m ($1.3$\%). Grid size ($10\times 10$)~m.  ]{\includegraphics[trim={0.0cm 0cm 0cm 0cm},clip,width=0.89\hsize]{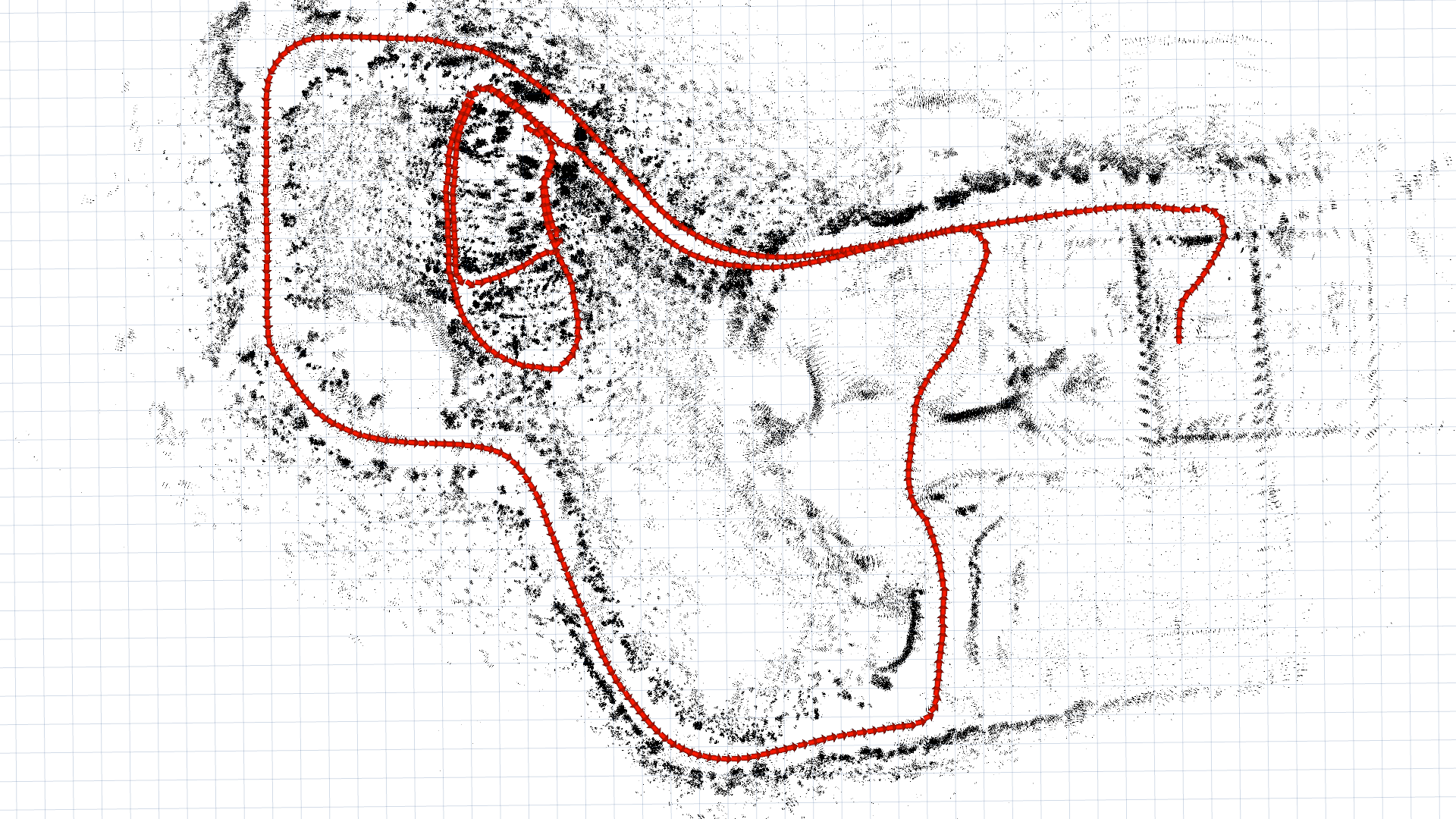}\label{fig:volvo_oxford_parameters}}\hfill 
    \\
    \subfloat[Odometry improved by tailoring parameters to the VolvoCE dataset (motion compensation switched off, $\zmin=85$, $f=3$). The Map is blurred around revisited regions of the large loop due to odometry drift but fairly sharp around the small loop. We estimate the maximum large loop position error to be $2.5$~m ($0.2\%$).]{\includegraphics[trim={0.0cm 0cm 0cm 0cm},clip,width=0.89\hsize]{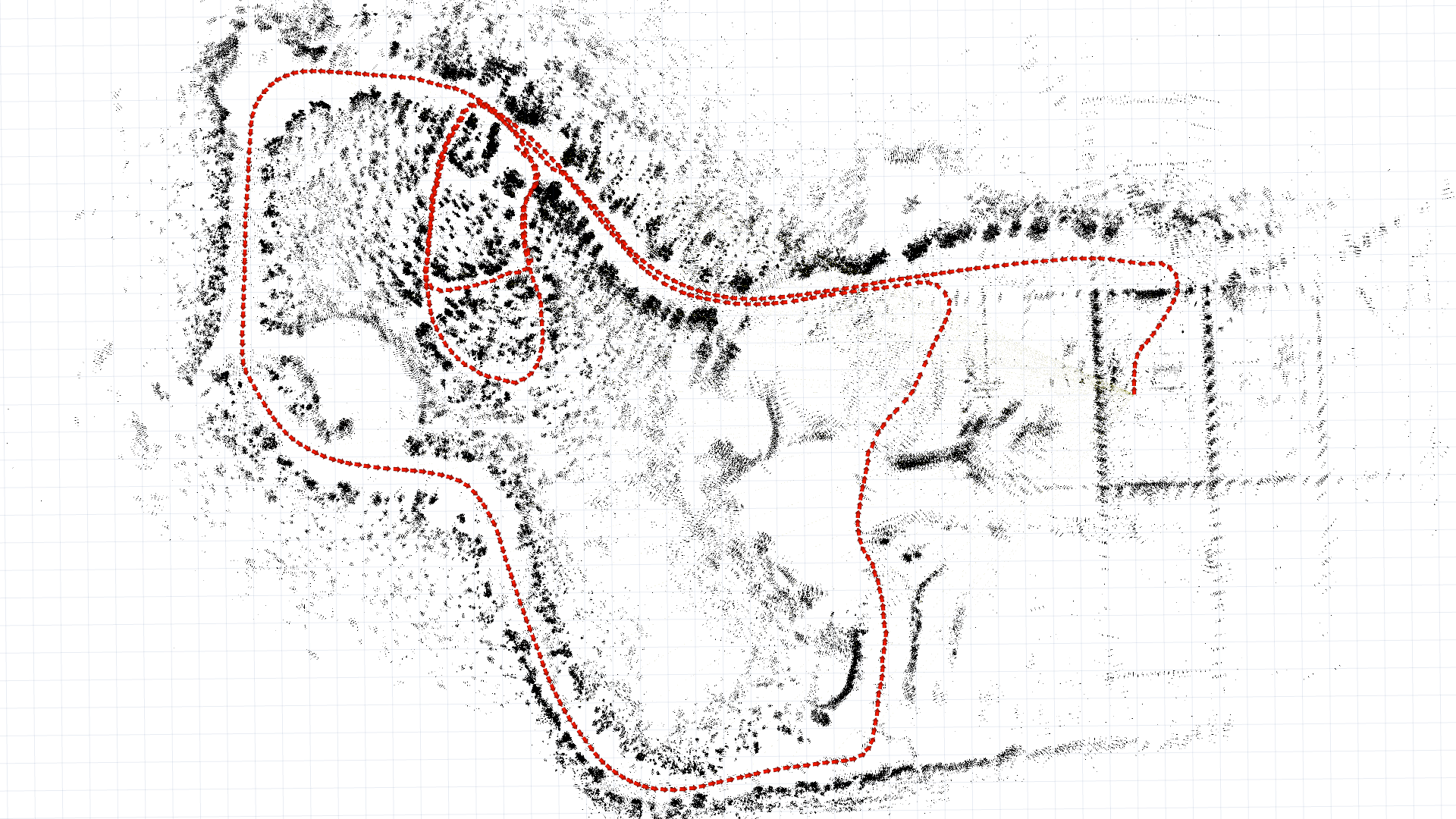}\label{fig:volvo_full_submap}}\hfill \\
    \subfloat[Full estimated trajectory with a single key-frame. The maximum large loop error is estimated to be $5$~m ($0.4\%$ error). The map around the large and small loop is more distorted compared to \ref{fig:volvo_full_submap}.]{\includegraphics[trim={0.0cm 0cm 0cm 0cm},clip,width=0.89\hsize]{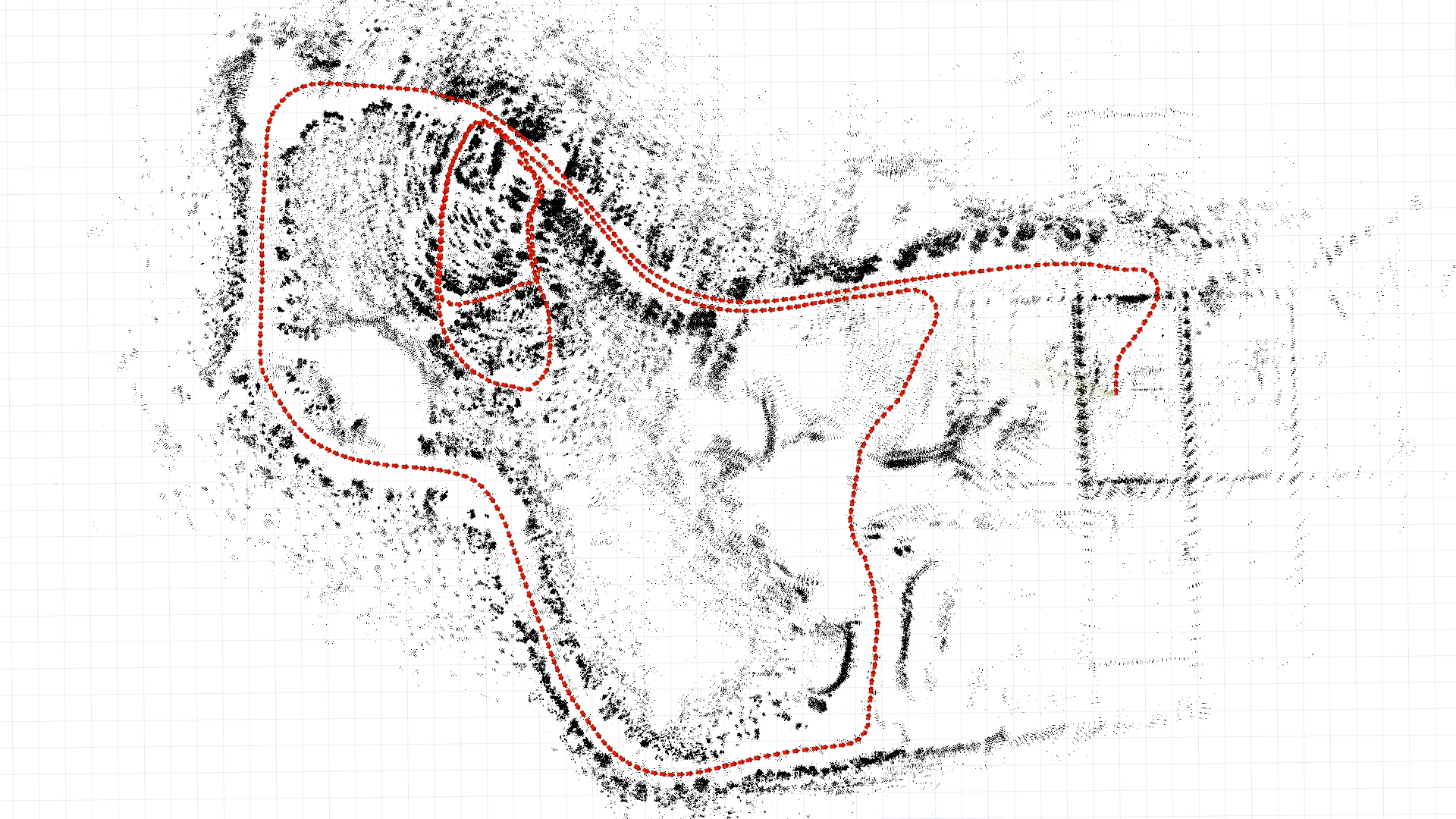}\label{fig:volvo_full_nosubmap}}\hfill \\
    \caption{Qualitative evaluation in VolvoCE dataset using various parameters.
    \label{fig:volvo_after_revisit}}
  \end{center}
  \vspace{-0.5cm}
\end{figure}

The estimated odometry and a corresponding map, obtained without changing parameters from the urban dataset, is depicted in Fig.~\ref{fig:volvo_oxford_parameters}. The maximum translation error is $15$~m ($1.3$\%) over the large loop. We notice that $\zmin=55$ is too low in combination with the sensor model and cluttered forest environment, however, \textit{k-strongest filtering} limits the amount of noise included in the filtered point cloud. We also note that motion compensation can have a negative impact in this dataset. For that reason, we made an attempt to improve the odometry by increasing the power threshold $\zmin=85$, by switching off motion compensation, and by increasing surface point density to $f=3$. These parameters reduced the estimated error to $2.5$~m ($0.2)$\% as depicted in Fig.~\ref{fig:volvo_full_submap}. For comparison, we plot the odometry with updated parameters but with a single submap keyframe $s=1$, this increases the error to $5~$m ($0.4$\%) as seen in Fig.~\ref{fig:volvo_full_nosubmap}.
In line with our previous results, increasing the number of submap keyframes reduces the drift. This can be seen by the relative increase in map blur in Fig.~\ref{fig:volvo_full_nosubmap} ($s = 1$) compared to Fig.~\ref{fig:volvo_full_submap} ($s = 3$). We believe this is because additionally constraining the registration reduces pose estimate noise and increases robustness due to uneven driving conditions, especially in the forest.


\subsubsection{Evaluation in mine -- Kvarntorp}
\label{sec:kvarntorp_eval}
In another experiment we mounted the Navtech CIR154XH radar on a car and drove through a $1235$-m long loop in an underground mine, at an approximate speed of $10$~km/h. The environment is depicted in Fig.~\ref{fig:kvarntorp_example}. The solid walls are highly radar-reflective and the interaction between radar, mountain walls and the car gives rise to a large amount of strong multi-path reflections. In contrast to the other evaluated datasets, the tunnels in parts of this environment have limited features that can constrain the registration in the longitudinal direction. 

Reusing the parameters tuned for Oxford dataset, we found that our method was able to reconstruct an accurate map of the environment with an estimated final position error of $15$~m ($1.2)$\% translation error. There are no visible errors in the passages between the top and bottom aisles in Fig.~\ref{fig:kvarntorp_example}, which is an indication that the method works well, even in a feature poor environment.


\section{Conclusions}
In this paper we have presented \ac{TITLE}, a radar odometry method that conservatively filters radar data using the $k$ strongest returns and represents the environment by a sparse set of oriented surface points. The oriented surface points can be used to efficiently and accurately perform scan registration by minimizing a point-to-line metric, obtaining outlier robustness using a Huber loss.
Our method is learning-free, requires no training data and can operate well in widely different environments and on different sensor models without changing a single parameter. Best odometry performance is, however, obtained by parameters tailored to the specific sensor and environment.
We found that drift can be further reduced by registering the latest scan to a history of key frames jointly, at the expense of computational time. Despite this, we were able to achieve 55~Hz single-thread performance on a laptop CPU, improving SCV state-of-the-art with $1.76$\% overall translation error in a public urban dataset.
Future work will focus on evaluating different loss functions and weighting residuals by surface uncertainty. 







\addtolength{\textheight}{-5cm}   
\bibliographystyle{IEEEtran}
\bibliography{references}

\begin{thebibliography}{10}
\providecommand{\url}[1]{#1}
\csname url@rmstyle\endcsname
\providecommand{\newblock}{\relax}
\providecommand{\bibinfo}[2]{#2}
\providecommand\BIBentrySTDinterwordspacing{\spaceskip=0pt\relax}
\providecommand\BIBentryALTinterwordstretchfactor{4}
\providecommand\BIBentryALTinterwordspacing{\spaceskip=\fontdimen2\font plus
\BIBentryALTinterwordstretchfactor\fontdimen3\font minus
  \fontdimen4\font\relax}
\providecommand\BIBforeignlanguage[2]{{%
\expandafter\ifx\csname l@#1\endcsname\relax
\typeout{** WARNING: IEEEtran.bst: No hyphenation pattern has been}%
\typeout{** loaded for the language `#1'. Using the pattern for}%
\typeout{** the default language instead.}%
\else
\language=\csname l@#1\endcsname
\fi
#2}}

\bibitem{todor_ndt}
T.~Stoyanov, M.~Magnusson, H.~Andreasson, and A.~Lilienthal, ``Fast and
  accurate scan registration through minimization of the distance between
  compact {3D NDT} representations,'' \emph{IJRR
  Robotics Research}, vol.~31, pp. 1377--1393, 09 2012.

\bibitem{gicp}
A.~Segal, D.~Hähnel, and S.~Thrun, ``Generalized-{ICP},'' 06 2009.

\bibitem{loam}
J.~Zhang and S.~Singh, ``Low-drift and real-time lidar odometry and mapping,''
  \emph{Autonomous Robots}, vol.~41, pp. 401--416, 02 2017.

\bibitem{behley2018rss}
J.~Behley and C.~Stachniss, ``Efficient surfel-based {SLAM} using {3D} laser
  range data in urban environments,'' in \emph{
  Systems~ (RSS)}, 2018.

\bibitem{zhu2017image}
J.~Zhu, ``Image gradient-based joint direct visual odometry for stereo
  camera.'' in \emph{IJCAI}, 2017, pp. 4558--4564.

\bibitem{Buczko_vis_odom}
M.~{Buczko} and V.~{Willert}, ``Flow-decoupled normalized reprojection error
  for visual odometry,'' in \emph{2016 IEEE 19th (ITSC)}, 2016, pp. 1161--1167.

\bibitem{6906584}
C.~{Forster}, M.~{Pizzoli}, and D.~{Scaramuzza}, ``{SVO}: Fast semi-direct
  monocular visual odometry,'' in \emph{2014 (ICRA)}, 2014, pp. 15--22.

\bibitem{7139486}
J.~{Zhang} and S.~{Singh}, ``Visual-lidar odometry and mapping: low-drift,
  robust, and fast,'' in \emph{2015 IEEE International Conference on Robotics
  and Automation (ICRA)}, 2015, pp. 2174--2181.

\bibitem{8202318}
H.~{Andreasson}, D.~{Adolfsson}, T.~{Stoyanov}, M.~{Magnusson}, and A.~J.
  {Lilienthal}, ``Incorporating ego-motion uncertainty estimates in range data
  registration,'' in \emph{2017 IEEE/RSJ (IROS)}, 2017, pp. 1389--1395.

\bibitem{narula_all-weather_2020}
L.~Narula, P.~A. Iannucci, and T.~E. Humphreys, ``Automotive-radar-based 50-cm
  urban positioning,'' in \emph{2020 IEEE/ION Position, Location and Navigation
  Symposium (PLANS)}, 2020, pp. 856--867.

\bibitem{acarballo2020libre}
A.~Carballo, J.~Lambert, A.~Monrroy, D.~Wong, P.~Narksri, Y.~Kitsukawa,
  E.~Takeuchi, S.~Kato, and K.~Takeda, ``{LIBRE}: The multiple 3d lidar
  dataset,'' \emph{arXiv preprint arXiv:2003.06129}, 2020, (accepted for
  presentation at IV2020).

\bibitem{barnes_masking_2020}
D.~Barnes, R.~Weston, and I.~Posner, ``Masking by moving: Learning
  distraction-free radar odometry from pose information,'' in \emph{CoRL}, ser.
  CoRL, L.~P. Kaelbling, D.~Kragic, and K.~Sugiura, Eds., vol. 100.\hskip 1em
  plus 0.5em minus 0.4em\relax PMLR, 30 Oct--01 Nov 2020, pp. 303--316.

\bibitem{barnes_under_2020}
D.~{Barnes} and I.~{Posner}, ``Under the radar: Learning to predict robust
  keypoints for odometry estimation and metric localisation in radar,'' in
  \emph{(ICRA)}, 2020, pp. 9484--9490.

\bibitem{burnett2021radar}
K.~Burnett, D.~J. Yoon, A.~P. Schoellig, and T.~D. Barfoot, ``Radar odometry
  combining probabilistic estimation and unsupervised feature learning,'' 2021.

\bibitem{hong2021radar}
Z.~Hong, Y.~Petillot, A.~Wallace, and S.~Wang, ``Radar slam: A robust slam
  system for all weather conditions,'' 2021.

\bibitem{8460687}
S.~H. {Cen} and P.~{Newman}, ``Precise ego-motion estimation with
  millimeter-wave radar under diverse and challenging conditions,'' in
  \emph{2018 IEEEn (ICRA)}, 2018, pp. 6045--6052.

\bibitem{8793990}
------, ``Radar-only ego-motion estimation in difficult settings via graph
  matching,'' in \emph{2019 (ICRA)}, 2019, pp. 298--304.

\bibitem{9197231}
Y.~S. {Park}, Y.~S. {Shin}, and A.~{Kim}, ``Pharao: Direct radar odometry using
  phase correlation,'' in \emph{2020 IEEE (ICRA)}, 2020, pp. 2617--2623.

\bibitem{vivet2013localization}
D.~Vivet, P.~Checchin, and R.~Chapuis, ``Localization and mapping using only a
  rotating fmcw radar sensor,'' \emph{Sensors}, vol.~13, no.~4, pp. 4527--4552,
  2013.

\bibitem{mielle-2019-comparative}
M.~Mielle, M.~Magnusson, and A.~J. Lilienthal, ``A comparative analysis of
  radar and lidar sensing for localization and mapping,'' in \emph{ecmr}, Sept.
  2019.

\bibitem{weston2019probably}
R.~Weston, S.~Cen, P.~Newman, and I.~Posner, ``Probably unknown: Deep inverse
  sensor modelling radar,'' in \emph{2019 (ICRA)}.\hskip 1em plus 0.5em minus
  0.4em\relax IEEE, 2019, pp. 5446--5452.

\bibitem{8794014}
R.~{Aldera}, D.~D. {Martini}, M.~{Gadd}, and P.~{Newman}, ``Fast radar motion
  estimation with a learnt focus of attention using weak supervision,'' in
  \emph{2019 (ICRA)}, 2019, pp. 1190--1196.

\bibitem{marck2013indoor}
J.~W. Marck, A.~Mohamoud, E.~vd~Houwen, and R.~van Heijster, ``Indoor radar
  slam a radar application for vision and gps denied environments,'' in
  \emph{2013 European Radar Conference}.\hskip 1em plus 0.5em minus 0.4em\relax
  IEEE, 2013, pp. 471--474.

\bibitem{Vivet_2013}
\BIBentryALTinterwordspacing
D.~Vivet, P.~Checchin, and R.~Chapuis, ``Localization and mapping using only a
  rotating {FMCW} radar sensor,'' \emph{Sensors}, vol.~13, no.~4, p.
  4527–4552, Apr 2013. [Online]. Available:
  \url{http://dx.doi.org/10.3390/s130404527}
\BIBentrySTDinterwordspacing

\bibitem{7795967}
F.~{Schuster}, C.~G. {Keller}, M.~{Rapp}, M.~{Haueis}, and C.~{Curio},
  ``Landmark based radar {SLAM} using graph optimization,'' in \emph{2016 IEEE
  19th International Conference on Intelligent Transportation Systems (ITSC)},
  2016, pp. 2559--2564.

\bibitem{8813841}
M.~{Holder}, S.~{Hellwig}, and H.~{Winner}, ``Real-time pose graph {SLAM} based
  on radar,'' in \emph{2019 IEEE Intelligent Vehicles Symposium (IV)}, 2019,
  pp. 1145--1151.

\bibitem{hong2020radarslam}
Z.~Hong, Y.~Petillot, and S.~Wang, ``Radarslam: Radar based large-scale slam in
  all weathers,'' in \emph{2020 (IROS)}, 2020, pp. 5164--5170.

\bibitem{burnett_we_2021}
K.~Burnett, A.~P. Schoellig, and T.~D. Barfoot, ``Do we need to compensate for
  motion distortion and doppler effects in spinning radar navigation?''
  \emph{IEEE RAL}, vol.~6, no.~2, pp. 771--778, 2021.

\bibitem{6907064}
D.~{Kellner}, M.~{Barjenbruch}, J.~{Klappstein}, J.~{Dickmann}, and
  K.~{Dietmayer}, ``Instantaneous ego-motion estimation using multiple
  {Doppler} radars,'' in \emph{2014 IEEE International Conference on Robotics
  and Automation (ICRA)}, 2014, pp. 1592--1597.

\bibitem{8917111}
R.~{Aldera}, D.~D. {Martini}, M.~{Gadd}, and P.~{Newman}, ``What could go
  wrong? introspective radar odometry in challenging environments,'' in
  \emph{2019 IEEE (ITSC)}, 2019, pp. 2835--2842.

\bibitem{saftescu_kidnapped_2020}
S.~Saftescu, M.~Gadd, D.~De~Martini, D.~Barnes, and P.~Newman, ``Kidnapped
  radar: Topological radar localisation using rotationally-invariant metric
  learning,'' in \emph{2020 IEEE International Conference on Robotics and
  Automation (ICRA)}, 2020, pp. 4358--4364.

\bibitem{gadd2020look}
M.~Gadd, D.~De~Martini, and P.~Newman, ``Look around you: Sequence-based radar
  place recognition with learned rotational invariance,'' in \emph{2020
  IEEE/ION Position, Location and Navigation Symposium (PLANS)}, 2020, pp.
  270--276.

\bibitem{8957240}
T.~Y. {Tang}, D.~{De Martini}, D.~{Barnes}, and P.~{Newman}, ``Rsl-net:
  Localising in satellite images from a radar on the ground,'' \emph{IEEE
  Robotics and Automation Letters}, vol.~5, no.~2, pp. 1087--1094, 2020.

\bibitem{tang2020selfsupervised}
T.~Y. Tang, D.~D. Martini, S.~Wu, and P.~Newman, ``Self-supervised localisation
  between range sensors and overhead imagery,'' 2020.

\bibitem{4543181}
A.~Censi, ``An icp variant using a point-to-line metric,'' in \emph{2008 IEEE
  ICRA}, 2008, pp. 19--25.

\bibitem{Huber1992}
\BIBentryALTinterwordspacing
P.~J. Huber, \emph{Robust Estimation of a Location Parameter}.\hskip 1em plus
  0.5em minus 0.4em\relax New York, NY: Springer New York, 1992, pp. 492--518.
  [Online]. Available: \url{https://doi.org/10.1007/978-1-4612-4380-9_35}
\BIBentrySTDinterwordspacing

\bibitem{ceres-solver}
\BIBentryALTinterwordspacing
S.~Agarwal, K.~Mierle, and Others, ``Ceres solver.'' [Online]. Available:
  \url{http://ceres-solver.org}
\BIBentrySTDinterwordspacing

\bibitem{RadarRobotCarDatasetICRA2020}
D.~Barnes, M.~Gadd, P.~Murcutt, P.~Newman, and I.~Posner, ``The oxford radar
  robotcar dataset: A radar extension to the oxford robotcar dataset,'' in
  \emph{2020 IEEE ICRA}, 2020, pp. 6433--6438.

\bibitem{Geiger2012CVPR}
A.~Geiger, P.~Lenz, and R.~Urtasun, ``Are we ready for autonomous driving? the
  {KITTI} vision benchmark suite,'' in \emph{
  Pattern Recognition (CVPR)}, 2012.

\bibitem{zhan2019dfvo}
H.~Zhan, C.~S. Weerasekera, J.~Bian, and I.~Reid, ``Visual odometry revisited:
  What should be learnt?'' \emph{arXiv preprint arXiv:1909.09803}, 2019.

\bibitem{lovelace_geocomputation_2019}
R.~Lovelace, J.~Nowosad, and J.~Muenchow, \emph{Geocomputation with
  {{R}}}.\hskip 1em plus 0.5em minus 0.4em\relax {CRC Press}, 2019.

\bibitem{Churchill2012ExperienceBN}
W.~S. Churchill, ``Experience based navigation : theory, practice and
  implementation,'' 2012.

\end{thebibliography}
\end{document}